\documentclass[runningheads]{llncs}

\usepackage{eccv}

\usepackage{eccvabbrv}

\usepackage{graphicx}
\usepackage{booktabs}

\usepackage{algorithm}
\usepackage{algpseudocode}
\usepackage{multirow}
\usepackage{booktabs}
\usepackage[table]{xcolor}
\usepackage{wrapfig}
\usepackage{subcaption}

\usepackage[accsupp]{axessibility}  

\usepackage{hyperref}

\usepackage{orcidlink}

\begin{document}

\title{SPICE: Simple Polysemantic Feature Interpretation via Clustering-based Explanation}

\titlerunning{SPICE: Simple Polysemantic Feature Interpretation}

\author{
Sehyun Lee\inst{1}\orcidlink{0009-0002-1711-9211} \and
Dahee Kwon\inst{1}\orcidlink{0009-0002-6395-8059} \and
Damin Lee\inst{1}\orcidlink{0009-0004-9380-1006} \and
Jaesik Choi\inst{1,2}\orcidlink{0000-0002-4663-3263}
}

\authorrunning{S.~Lee et al.}

\institute{
Korea Advanced Institute of Science and Technology (KAIST), Daejeon, Republic of Korea\\
\email{\{sehyun.lee, daheekwon, jaesik.choi\}@kaist.ac.kr, xxdamin@gmail.com}
\and
INEEJI, Seongnam, Republic of Korea
}

\maketitle

\begin{abstract}
  One of the pivotal recent challenges in neural network interpretability is polysemanticity, where a single neuron is activated by multiple, often unrelated concepts, hindering clear functional understanding. Although prior work has explored this phenomenon, existing approaches remain architecture-specific and depend on manual heuristics such as a fixed number of concept clusters ($K$), limiting their generality and scalability—especially for modern Transformer-based models. To address these limitations, we introduce SPICE (\textbf{S}imple \textbf{P}olysemantic Feature \textbf{I}nterpretation via \textbf{C}lustering-based \textbf{E}xplanation), a generalizable framework for analyzing polysemanticity in deep vision architectures. SPICE avoids architecture-dependent propagation rules, enabling the first systematic comparison of polysemanticity across both CNNs and Transformers, and automatically determines the number of concept clusters per neuron, eliminating reliance on a preset $K$ and supporting scalable analysis for large models. Using SPICE, we conduct a comprehensive investigation into how polysemanticity emerges, varies across depth and architecture, and forms through distinct computational pathways.
\end{abstract}

\section{Introduction}
\label{sec:intro}

Deep learning models have achieved remarkable performance across a wide range of applications, but their rapidly increasing complexity and scale have made it ever more difficult to understand their underlying mechanisms~\cite{arrieta2020explainable}. As a result, interpretability research—which seeks to clarify the internal representations and computational processes learned by models—has become increasingly important and is now an active area of study. In particular, mechanistic interpretability, which examines the functions of neurons and modules in storing and processing knowledge, has emerged as one of the most prominent research directions in the field~\cite{olah2017feature,olah2020zoom, bereska2024mechanistic, elhage2021mathematical}.

In the vision domain, such research has been actively pursued, with approaches such as those in \cite{bau2017network, OikarinenW23CLIPdissect} analyzing models at the level of individual neurons by identifying shared visual features among the samples that strongly activate them. This approach rests on the implicit assumption that each neuron corresponds to a single function. Recent studies, however, have shown that multiple concepts often coexist within a single neuron—a phenomenon known as \textbf{\textit{polysemanticity}}~ \cite{elhage2022toy, dreyer2024pure}. As a result, single-neuron analysis faces fundamental limitations in faithfulness, and such interpretations alone are insufficient to fully explain the underlying mechanisms of deep models.

To account for polysemanticity, several approaches have been proposed. While training interpretable Sparse Autoencoders (SAEs) offers one promising path, such methods do not directly reveal the intrinsic structural flow of the network~\cite{cunningham2023sparse,thasarathan2025universal}. To address this, attribution clustering has emerged as a key technique for operationalizing circuit-based tracing by grouping a neuron's attribution footprints to separate its encoded concepts. However, existing attribution clustering methods suffer from two critical limitations. First, they exhibit a fundamental lack of generality. To date, these analyses have been almost exclusively confined to CNNs, failing to provide a unified framework applicable to modern Transformer-based architectures. Second, they lack scalability due to their reliance on manual heuristics. Specifically, these approaches often require pre-defining a fixed number of concept clusters ($K$) for each neuron-a subjective and laborious process that is infeasible for today's large-scale models. Consequently, developing a generalizable and scalable method to analyze polysemanticity across diverse architectures remains a key unsolved problem.

\begin{figure}[t]
\begin{center}
\includegraphics[width=\textwidth]{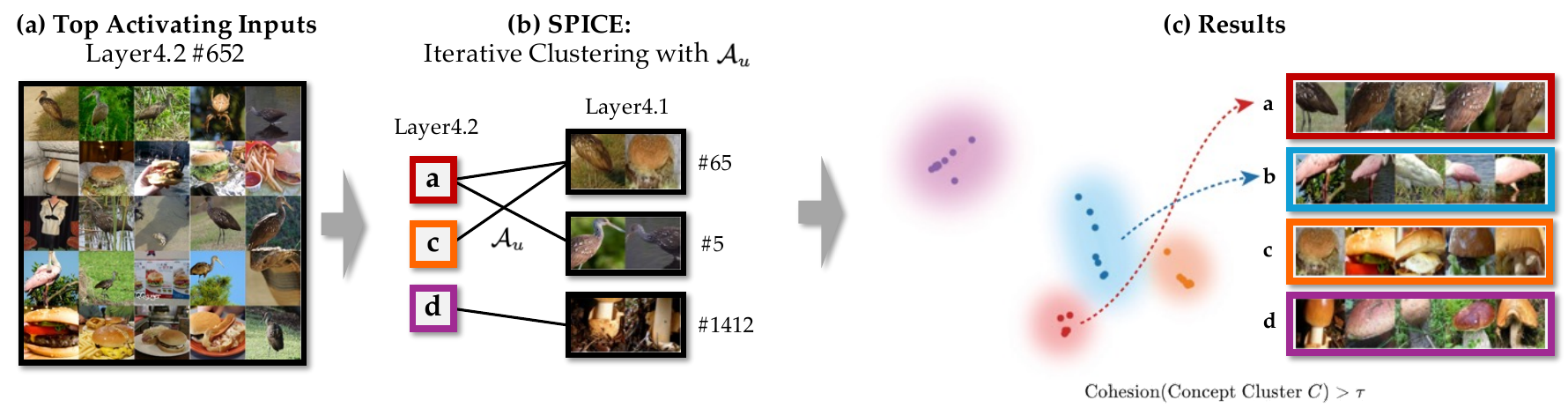}
\end{center}
\caption{\textbf{An overview of the proposed method (SPICE).} \textbf{(a)} A polysemantic neuron exhibits mixed selectivity across diverse image inputs. \textbf{(b)} SPICE traces the computational pathways of these activations by computing attribution footprints ($\mathcal{A}_u$) from the preceding layer. \textbf{(c)} We iteratively cluster these footprints to disentangle the neuron's representations. By strictly separating clusters that satisfy the condition $\text{Cohesion}(C) > \tau$, SPICE successfully decomposes the entangled inputs into highly coherent, interpretable conceptual groups.}
\label{fig:overview}
\end{figure}

Therefore, we introduce SPICE (\textbf{S}imple \textbf{P}olysemantic Feature \textbf{I}nterpretation via \textbf{C}lustering-based \textbf{E}xplanation), a method that disentangles polysemantic neurons with an emphasis on generality and scalability. SPICE is both simple and effective, addressing two major limitations of prior work.
(1) We introduce a scalable and flexible attribution-clustering framework for dissecting polysemantic neurons. By adaptively determining the optimal number of concept clusters ($K$), SPICE eliminates the reliance on manual heuristics. Furthermore, extensive evaluations and ablations demonstrate that our method achieves highly robust disentanglement performance across diverse architectures, extreme activation ranges, and various attribution methods. (2) Using this framework, we provide an in-depth analysis of polysemanticity to explore the representational differences across models. Our findings highlight distinct inductive biases between CNNs and Transformers, and illustrate the diverse computational pathways that underlie concept formation.

Using the proposed method, we conduct an in-depth investigation into the formation patterns of polysemanticity and their internal structure. Beyond extending interpretability, this analysis offers novel insights into the fundamental principles by which models organize and represent concepts.
\section{Related Work}

\paragraph{Mechanistic Interpretability}
A significant branch of interpretability research focuses on identifying human-understandable concepts and mechanisms within neural networks~\cite{bereska2024mechanistic}. This line of work seeks to move beyond input–output correlations and explain how a model computes internally. Early studies focused on associating individual neurons with specific visual concepts~\cite{bau2017network, olah2017feature, olah2020zoom}. Since then, research has broadened to include methods for automatically discovering concepts with limited human supervision~\cite{kalibhat2023identifying, OikarinenW23CLIPdissect}, as well as approaches that analyze circuits—pathways of functionally related neurons—to uncover higher-level computational structures~\cite{NeurFlow, wang2022interpretability, VCC, kwon2025granular}. Our work builds on this tradition, aiming to provide a scalable and systematic analysis of a neuron's internal conceptual structure.

\paragraph{Polysemanticity as a Challenge}
Early interpretability methods often relied on an implicit monosemantic assumption—that each neuron corresponds to a single coherent concept. However, polysemanticity is essentially a fundamental property of deep networks, arising from data statistics and superposition constraints~\cite{marshall2024understanding, elhage2022toy}. To account for this, Sparse Autoencoders (SAEs) are used to disentangle compressed features~\cite{cunningham2023sparse, thasarathan2025universal, fel2025archetypal}, though they do not directly reveal the model’s intrinsic attribution flow. Furthermore, while several works leverage Vision-Language Models (VLMs) to probe polysemantic concepts~\cite{achiam2023gpt, zhai2023SigLIP, oikarinen2024linear, yu2025coe}, these approaches are fundamentally bottlenecked by representation misalignment and the sensitivity to the chosen external model.

\paragraph{Disentanglement via Attribution Approach}
To directly analyze the internal structure of polysemantic neurons, several approaches leverage attribution clustering—grouping a neuron's attribution footprints to separate its encoded concepts~\cite{dreyer2024pure, hesse2025disentangling, yu2025coe}. These methods are powerful, but the literature exhibits two critical limitations that our work addresses. First, a lack of generality across architectures. To date, these analyses~\cite{dreyer2024pure, hesse2025disentangling} have been almost exclusively confined to CNN architectures. A unified framework to systematically analyze and compare polysemanticity in modern Transformer-based models has been notably absent. Second, prior work lacks scalability in concept discovery, largely due to reliance on manual or restrictive heuristics. Pioneering work like PURE~\cite{dreyer2024pure} requires manually specifying the number of concepts ($K$) for each neuron, while methods such as~\cite{hesse2025disentangling} perform explicit disentanglement but are constrained to separating only a small, fixed number of components rather than discovering an arbitrary, data-driven set of clusters. Both approaches are subjective, time-consuming, and practically infeasible for large-scale analysis. Therefore, a framework that is both generalizable and scalable for attribution-based analysis has remained a key unsolved problem, which we address in this paper.
\section{Method}
\label{sec:method}

The visual concept captured by a single neuron is often inferred from the shared features of the samples that most strongly activate it \cite{bau2017network, kalibhat2023identifying, OikarinenW23CLIPdissect}. While this approach can reveal what a neuron has learned, it becomes challenging to disentangle multiple concepts when a neuron is polysemantic—i.e., when several distinct concepts coexist within the same unit. To analyze this polysemanticity within a model’s learned visual representations, our goal is therefore to cluster the multiple visual concepts encoded by a single neuron. To this end, we introduce \textbf{S}imple \textbf{P}olysemantic Feature \textbf{I}nterpretation via \textbf{C}lustering-based \textbf{E}xplanation (SPICE).

\paragraph{Identifying Representational Footprints via Attribution}
Information in deep neural networks is widely understood to be formed hierarchically, beginning at the input and propagating through successive layers until it culminates in the final decision~\cite{zeiler2014visualizing, lecun2015deep, olah2020zoom}. Consequently, understanding an internal visual concept requires tracing its origins along this information flow. Within this framework, we seek to explain the information encoded by each neuron by defining its representational footprint-a trace of its compositional origins. Building upon this premise, we assume that concepts sharing similar footprints are treated as `related' by the model. To realize this empirically, we compute these footprints using attribution methods, which are widely employed for analyzing deep vision models~\cite{shrikumar2017learning,selvaraju2017grad, simonyan2014visualising}. 

Formally, given a dataset \(\mathcal{D}=\{x_i\}\) and a network \(f\), we first interpret the concept represented by a neuron \(u^l\) in layer \(l\) through its set of most highly activating samples, \(\mathcal{D}_{u^l}\subset\mathcal{D}\). We define a \emph{neuron} according to the network architecture. For convolutional models, each channel is treated as a neuron, as it acts as a distinct feature detector across spatial locations.\footnote{The term ``neuron'' is often used interchangeably with ``feature'' in the literature, or ``channel'' in the context of CNNs.} For transformer-based models, each hidden dimension is treated as a neuron, as it represents a distinct feature across all input patches.

To trace the origin of the visual concept represented by a single neuron \(u^l\), we quantify how each predecessor neuron in layer \(l-1\) contributes to its activation. Since a neuron corresponds to an entire channel in CNNs or a hidden dimension in Transformers, its activation is distributed over multiple output positions, i.e., a spatial feature map in CNNs or a sequence of patch activations in Transformers. We therefore aggregate the positional activations to obtain a scalar neuron activation,

\begin{equation}
\bar a_u(x):=
\begin{cases}
\sum_{h,w} a_u(x)_{h,w}, & \text{(CNN)},\\
\sum_{p} a_u(x)_p, & \text{(Transformer)}.
\end{cases}
\end{equation}

For a given sample \(x_i\), the aggregated activations of all predecessor neurons form the activation vector

\begin{equation}
\bar{\mathbf a}^{\,l-1}(x_i)
=
\left[
\bar a_{u_1^{l-1}}(x_i),
\ldots,
\bar a_{u_{d_{l-1}}^{\,l-1}}(x_i)
\right]
\in
\mathbb R^{d_{l-1}}.
\end{equation}

Using the aggregated activation of the target neuron as the scalar objective, we compute the Input\(\times\)Gradient attribution from all predecessor neurons to target neuron,

\begin{equation}
\boldsymbol{\phi}(x_i)
=
\bar{\mathbf a}^{\,l-1}(x_i)
\odot
\frac{\partial \bar a_{u^l}(x_i)}
{\partial \bar{\mathbf a}^{\,l-1}(x_i)}
\in
\mathbb R^{d_{l-1}},
\end{equation}

where \(\odot\) denotes element-wise multiplication. The \(j\)-th entry of \(\boldsymbol{\phi}(x_i)\) represents the attribution from predecessor neuron \(u_j^{l-1}\) to the target neuron \(u^l\).

Stacking the attribution vectors of all highly activating samples \(x_i\in\mathcal D_{u^l}\) yields the attribution footprint matrix,

\begin{equation}
\mathcal A_{u^l}
=
\begin{bmatrix}
\boldsymbol{\phi}(x_1)^\top\\
\boldsymbol{\phi}(x_2)^\top\\
\vdots\\
\boldsymbol{\phi}(x_{|\mathcal D_{u^l}|})^\top
\end{bmatrix}
\in
\mathbb R^{|\mathcal D_{u^l}|\times d_{l-1}}.
\end{equation}

Clustering the rows of \(\mathcal A_{u^l}\) therefore groups samples with similar predecessor-attribution patterns, revealing distinct concepts that drive the activation of \(u^l\).

As model depth increases, the attribution vectors of different highly activating samples of the same neuron (i.e., the rows of \(\mathcal{A}_u\)
) become substantially more similar to one another. We hypothesize that this effect stems from the increasing anisotropy of the feature space, a phenomenon widely reported in prior work on large models~\cite{godey2024anisotropy, razzhigaev2024shape, bansal2018can, elhage2023privileged}. To mitigate its impact and enable effective attribution footprint-based clustering, we normalize attribution matrices along the neuron dimension before clustering. Rather than simply placing dimensions on a comparable scale, this normalization actively mitigates the anisotropy problem, allowing the clustering algorithm to focus on relative semantic patterns rather than absolute magnitude biases. With this normalization step, our method achieves more reliable concept disentanglement. 

\paragraph{Disentangling Concepts by Clustering Footprints}
Having defined the representational footprints, we now introduce our method, namely SPICE. Rather than assuming a fixed number of underlying concepts, we propose an iterative algorithm that progressively identifies and separates cohesive groups of footprints based on their pairwise cosine similarity. As detailed in Algorithm~\ref{alg:disentangling}, the process begins by attempting to partition the rows of the footprint matrix \(\mathcal{A}_u\) into \(k=2\) clusters. A cluster \(\mathcal{C}\)  is considered a cohesive concept if its internal coherence, which we define as the average pairwise cosine similarity between its members, exceeds an adaptive threshold \(\tau\). Cohesive concepts are accepted as final and removed from the working set \(\mathcal{A}'\). The algorithm then dynamically adjusts the number of clusters \(k\) and repeats this process on the remaining attributions until no further cohesive concepts can be disentangled. 

A key component is the dynamic calculation of this cohesion threshold \(\tau\). A fixed, universal threshold would fail to adapt to the diverse similarity distributions exhibited by the footprints of different neurons. We therefore compute \(\tau\) adaptively from the set of footprints currently under analysis, \(\mathcal{A}'\). Our empirical analysis reveals that the character of the similarity distribution systematically changes with network depth, as illustrated in Figure~\ref{fig:basic_stat}(right). Early layers typically exhibit a unimodal, long-tailed distribution of similarities. In contrast, later layers often display a bimodal distribution, where the emergence of a second peak suggests the formation of distinct, highly cohesive conceptual groups.

To create a robust criterion that accommodates both scenarios, we define the threshold \(\tau\) as the maximum of two candidates: (1) the 95th percentile of the pairwise similarity values, and (2) the value corresponding to the second peak of a Kernel Density Estimate (KDE) fitted to the similarity distribution. Formally, letting \(S\) denote the set of pairwise cosine similarities of the current footprints \(\mathcal{A}'\),
\begin{equation}
\tau = \max\bigl(Q_{0.95}(S),\, p_2(S)\bigr),
\end{equation}
where \(Q_{0.95}(S)\) is the 95th percentile of \(S\) and \(p_2(S)\) is the second local maximum (in order of increasing similarity) of the Gaussian KDE of \(S\) whose PDF value exceeds \(0.1\); when the distribution is unimodal and \(p_2(S)\) does not exist, \(\tau\) falls back to \(Q_{0.95}(S)\). The percentile provides a robust baseline, particularly for the unimodal distributions in early layers, while the second peak offers a more semantically meaningful partition for the well-formed concepts in later layers. This dual-criterion approach ensures that our definition of a concept is contextually grounded across the network's hierarchy. 

\begin{algorithm}[!t]
\caption{SPICE}
\label{alg:disentangling}
\begin{algorithmic}[1]
\State \textbf{Input:} A set of attribution footprints $\mathcal{A}_u$, a cohesion threshold $\tau$.
\State \textbf{Output:} A set of disjoint concepts $\mathcal{C}^*$, where each concept is a set of footprints.
\State
\Procedure{DisentangleConcepts}{$\mathcal{A}_u, \tau$}
    \State \textcolor{blue}{$\triangleright$  \textit{Initialize working variables}}
    \State $\mathcal{A}' \leftarrow \mathcal{A}_u$
    \State $\mathcal{C}^* \leftarrow \emptyset$
    \State $k \leftarrow 2$

    \While{$|\mathcal{A}'| \ge k$}
        \State \textcolor{blue}{$\triangleright$ \textit{Search for cohesive clusters of size $k$}}
        \State Let $\mathcal{C}_{cohesive}$ be the set of all clusters $C \in \text{Cluster}(\mathcal{A}', k)$ where $\text{Cohesion}(C) > \tau$.

        \If{$\mathcal{C}_{cohesive} \neq \emptyset$}
            \State \textcolor{blue}{$\triangleright$ \textit{Found cohesive clusters: update}}
            \State $\mathcal{C}^* \leftarrow \mathcal{C}^* \cup \mathcal{C}_{cohesive}$
            \State $\mathcal{A}' \leftarrow \mathcal{A}' \setminus \bigcup_{C \in \mathcal{C}_{cohesive}} C$
            \State $k \leftarrow \max(2, k - (|\mathcal{C}_{cohesive}| - 1))$
        \Else
            \State \textcolor{blue}{$\triangleright$ \textit{No cohesive cluster: increase search size}}
            \State $k \leftarrow k + 1$
        \EndIf
    \EndWhile

    \State \textcolor{blue}{$\triangleright$ \textit{Treat remaining footprints as individual concepts}}
    \State $\mathcal{C}^* \leftarrow \mathcal{C}^* \cup \{ \{a\} \mid a \in \mathcal{A}' \}$
    \State \Return $\mathcal{C}^*$
\EndProcedure

\Statex
\Statex where $\text{Cohesion}(C) = \text{mean}_{a_i, a_j \in C} (\text{sim}(a_i, a_j))$
\end{algorithmic}
\end{algorithm}
\section{Experiment}
In this section, we present both the qualitative and quantitative evaluation of our proposed method, along with a comprehensive analysis. 

\subsection{Quantitative Comparison with Baselines}

\begin{table}[t]
\newcommand{\res}[2]{$#1{\tiny_{\pm #2}}$}
\caption{Separability comparison of baselines versus \textbf{Ours} across models and layers. As PURE was originally designed for CNNs, we adapted its CRP implementation for Transformer models for this comparison.}
\centering
\small
\setlength{\tabcolsep}{4pt}

\begin{tabular}{l|ccc>{\columncolor{gray!20}}c}
\toprule
\textbf{Model / Layer} & \textbf{PURE}$^*$ & \textbf{LE} & \textbf{CPE} & \textbf{Ours} \\
\midrule

ResNet-50 L2.1
& \res{1.018}{0.031} & \res{\textbf{1.112}}{0.122} & \res{0.996}{0.014} & \res{1.092}{0.056} \\

ResNet-50 L3.3 
& \res{0.999}{0.012} & \res{1.108}{0.153} & \res{0.995}{0.007} & \res{\textbf{1.160}}{0.081} \\

ResNet-50 L4.2 
& \res{1.009}{0.071} & \res{1.187}{0.221} & \res{0.998}{0.011} & \res{\textbf{1.291}}{0.224} \\

\midrule

ViT-B B2 
& \res{1.002}{0.010} & \res{\textbf{1.199}}{0.181} & \res{0.994}{0.009} & \res{1.137}{0.062} \\

ViT-B B6 
& \res{1.012}{0.013} & \res{1.185}{0.153} & \res{0.994}{0.009} & \res{\textbf{1.240}}{0.089} \\

ViT-B B11 
& \res{1.035}{0.025} & \res{1.357}{0.181} & \res{0.998}{0.018} & \res{\textbf{1.577}}{0.089} \\

\midrule

DenseNet L2.1
& \res{1.012}{0.028} & \res{\textbf{1.090}}{0.158} & \res{0.994}{0.007} & \res{1.071}{0.059} \\

DenseNet L3.12 
& \res{1.053}{0.054} & \res{1.062}{0.139} & \res{0.994}{0.005} & \res{\textbf{1.114}}{0.060} \\

DenseNet L4.16 
& \res{1.025}{0.035} & \res{\textbf{1.191}}{0.116} & \res{0.994}{0.009} & \res{1.033}{0.032} \\

\midrule

CLIP ViT-B B2 
& \res{1.007}{0.018} & \res{1.088}{0.082} & \res{0.994}{0.008} & \res{\textbf{1.090}}{0.051} \\

CLIP ViT-B B6 
& \res{1.008}{0.016} & \res{1.165}{0.148} & \res{0.997}{0.009} & \res{\textbf{1.212}}{0.090} \\

CLIP ViT-B B11 
& \res{1.021}{0.040} & \res{1.244}{0.192} & \res{0.998}{0.013} & \res{\textbf{1.410}}{0.102} \\

\bottomrule
\end{tabular}
\label{tab:intra_inter_clustering_merged}
\end{table}

To demonstrate the effectiveness of our method in disentangling and explaining multiple concepts encoded within a single neuron, we extensively evaluate SPICE across four diverse architectures (ResNet-50, ViT-B, DenseNet, and CLIP ViT-B). We compare our approach against several recent baselines, including PURE~\cite{dreyer2024pure}, CPE~\cite{yu2025coe}, and LE~\cite{oikarinen2024linear}. While PURE was originally restricted to CNNs due to its reliance on CRP grammar, we extended and adapted its implementation for Transformer architectures to ensure a comprehensive evaluation. Moreover, unlike SPICE-which adaptively determines the optimal number of clusters per neuron-PURE requires the number of clusters to be specified in advance. Accordingly, we configure it by selecting the optimal number of clusters using the silhouette score.

\subsubsection{Separability based on External Semantic Space.}
To quantitatively evaluate the effectiveness of our concept disentanglement, we first adopt \textit{separability} as our primary metric, defined as the ratio of intra-cluster similarity to inter-cluster similarity. Following the established protocol of PURE, we utilize CLIP as the external verification tool to measure these similarities, grounded in the premise that visually similar images cluster closely in the hidden space. Concretely, for each neuron $n$ with discovered clusters, we compute $\bar{s}^{(n)}_{\mathrm{intra}}$ as the mean CLIP cosine similarity over all within-cluster image pairs and $\bar{s}^{(n)}_{\mathrm{inter}}$ as the mean over \emph{all} cross-cluster pairs (i.e., the full $i\!\times\!j$ pairwise comparisons, not centroid-based), and report a single dataset-level ratio of means,
\begin{equation}
\mathrm{Sep.} = \frac{\frac{1}{N}\sum_{n=1}^{N}\bar{s}^{(n)}_{\mathrm{intra}}}{\frac{1}{N}\sum_{n=1}^{N}\bar{s}^{(n)}_{\mathrm{inter}}}.
\end{equation}
We use the ImageNet validation set as the probing dataset for all separability evaluations. A higher separability score indicates a more successful and distinct concept separation.

As shown in Table~\ref{tab:intra_inter_clustering_merged}, SPICE exhibits dominant separability scores across various networks and depths. While text-prior-based methods like LE show competitive performance in early layers (e.g., B2), their performance notably degrades in deeper layers (e.g., L4.2, B11) where concepts become highly abstract and polysemantic. In contrast, SPICE maintains robust disentanglement, significantly outperforming all baselines in the deep layers. The slight deviation in early layers occurs because they typically capture low-level features (e.g., textures, colors) that are less aligned with CLIP's high-level semantic space. We note that CPE consistently yields near-unit separability ($\approx 0.99$); this stems from an evaluation mismatch rather than a low cluster count: CPE groups samples by VLM-generated text concepts whereas separability is measured in CLIP image-embedding space, and under the CPE protocol each sample is linked to roughly three concepts per neuron so the same images recur across groups, pulling inter-cluster similarity toward intra-cluster similarity. This is further evidenced by PURE, which uses an even smaller fixed $K\!=\!2$ yet attains a higher score (e.g., 1.035 at ViT-B B11), ruling out cluster count as the cause. This phenomenon is visually supported in Figure~\ref{fig:comparison}.

\begin{figure}[t]
    \centering
    \includegraphics[width=0.95\linewidth]{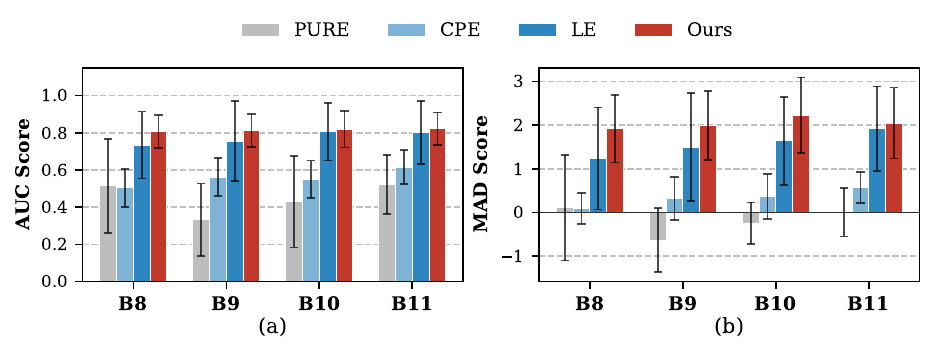}
    \caption{Simulation-based Semantic Evaluation in ViT-B}
    \label{fig:cosy_eval}
\end{figure}

\subsubsection{Simulation-based Evaluation.}
To further rigorously validate the semantic faithfulness of the discovered concepts beyond embedding-based metrics, we conduct a generative evaluation using the CoSy protocol~\cite{kopf2024cosy}. Specifically, we generate one textual description per discovered cluster with GPT-4.1~\cite{achiam2023gpt} and synthesize verification images via SDXL, following the CoSy protocol. AUC (Alignment Under Curve) and MAD (Mean Absolute Difference) are computed per cluster and then averaged, so that the resulting score is independent of the number of clusters $K$ each method assigns to a neuron.

As illustrated in Figure~\ref{fig:cosy_eval}, SPICE consistently achieves the highest AUC and MAD scores across multiple blocks of ViT-B compared to PURE, CPE, and LE. It is worth noting that while inherent text-visual misalignment exists in deep vision models, SPICE achieves higher validity because it intrinsically discovers concept clusters from internal activations independent of text. By doing so, it effectively bypasses the bottleneck of text-defined priors that fundamentally limits baselines like LE and CPE, ensuring a more faithful recovery of the model's actual learned concepts.

\subsection{Qualitative Comparison with Baselines}
To visually demonstrate the effectiveness of our method in disentangling and explaining multiple concepts encoded within a single neuron, we perform a qualitative comparison against the aforementioned baselines (PURE~\cite{dreyer2024pure}, CPE~\cite{yu2025coe}, and LE~\cite{oikarinen2024linear}).

\begin{figure}[ht!]
\begin{center}
    \includegraphics[width=\textwidth]{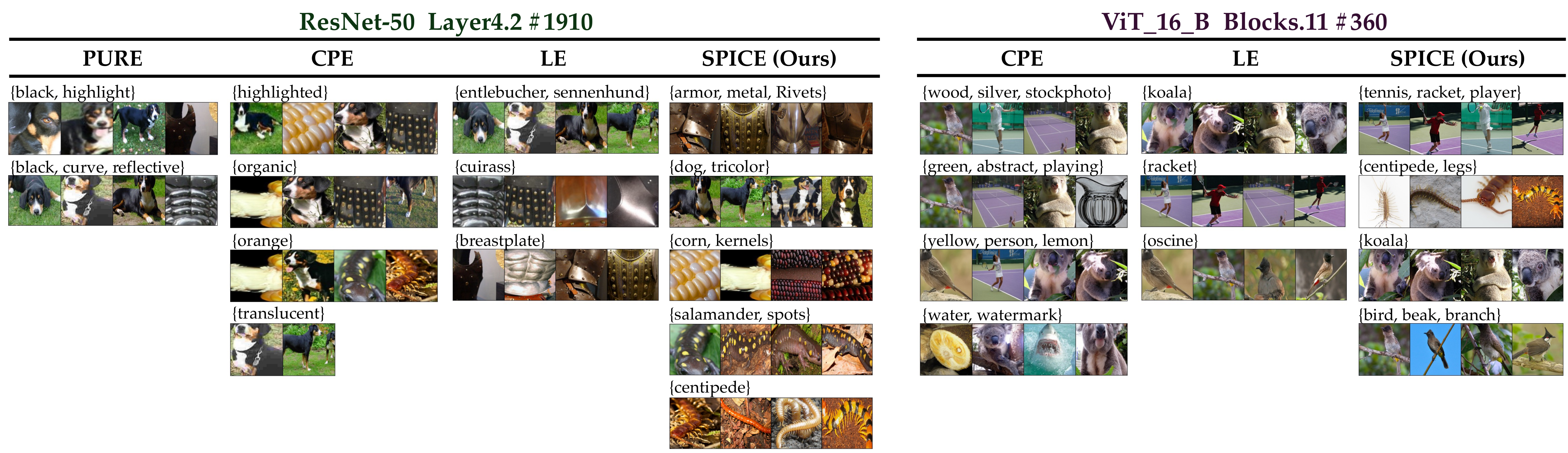}
\end{center}
\caption{Qualitative comparisons with various baselines. Each row shows four exemplar images representing the visual characteristics of a discovered cluster, accompanied by its corresponding textual description.}
\label{fig:comparison}
\end{figure}

In Figure~\ref{fig:comparison}, each row illustrates representative images belonging to a single concept cluster discovered by each method. Across both ResNet-50 and ViT-B/16, SPICE discovers the most comprehensive set of polysemantic clusters, indicating its strong capability to fully recover the diverse concepts encoded within a neuron. Moreover, while several baselines exhibit notable visual inconsistency and noise within their clusters, SPICE consistently yields semantically aligned and coherent groupings. Similar to LE and CPE, SPICE naturally integrates with vision-language models~\cite{achiam2023gpt} to provide accurate textual descriptions for each cluster. Taken together, these qualitative results demonstrate that SPICE offers the most reliable visual characterization of neuron-level polysemanticity.

\subsection{In-depth Analysis}
Using our framework, we conduct a comprehensive analysis of polysemanticity. We first visually dissect a single neuron, and then trace the computational pathways that give rise to these diverse concepts. We then broaden our scope to explore architectural inductive biases across models, and finally validate the robustness of our findings through rigorous ablation studies.

\begin{figure}[b]
    \centering
    \begin{subfigure}[t]{0.24\textwidth}
        \centering
        \includegraphics[width=0.9\textwidth]{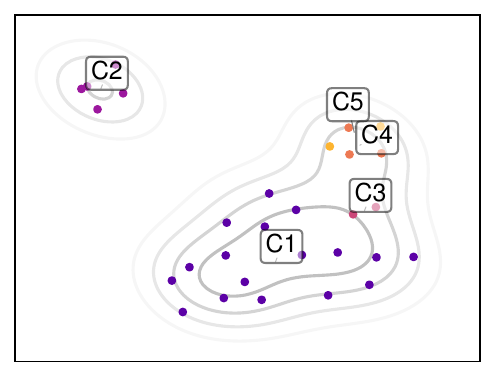} 
        \caption{}
        \label{fig:umap_projection}
    \end{subfigure}
    \hfill
    \begin{subfigure}[t]{0.24\textwidth}
        \centering
        \includegraphics[width=0.9\textwidth]{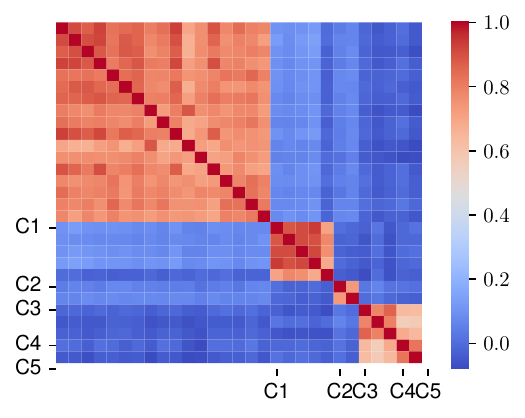}
        \caption{}
        \label{fig:similarity_heatmap}
    \end{subfigure}
    \hfill
    \begin{subfigure}[t]{0.22\textwidth} 
        \centering
        \includegraphics[width=\textwidth]{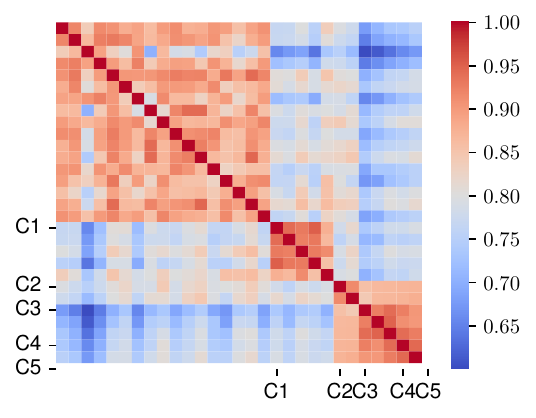} 
        \caption{}
        \label{fig:clip_heatmap}
    \end{subfigure}
    \hfill
    \begin{subfigure}[t]{0.25\textwidth}
        \centering
        \includegraphics[width=\textwidth]{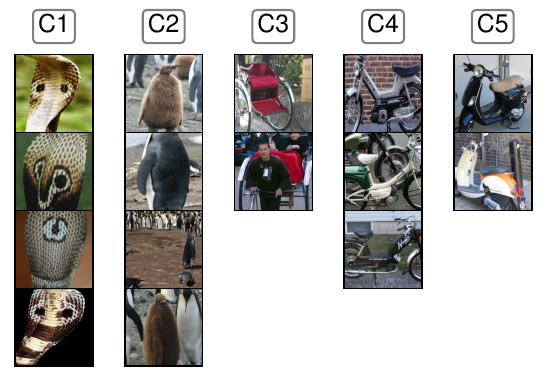} 
        \caption{}
        \label{fig:exemplar_images}
    \end{subfigure}

    \caption{Dissecting a polysemantic neuron (\#652) from ViT-B block 11. \textbf{(a)} UMAP embedding reveals distinct conceptual clusters within the neuron's representation footprints. \textbf{(b)} A similarity heatmap quantifies the low inter-cluster and high intra-cluster similarity. \textbf{(c)} A CLIP similarity heatmap quantitatively shows that concepts are semantically disparate. \textbf{(d)} Exemplar images confirm the diverse and disparate visual concepts represented by each cluster; such as snakeskin (C1), penguins (C2), and various two-wheeled vehicles (C3-C5).}
    \label{fig:polysemantic_neuron_dissection}
\end{figure}

\subsubsection{Dissecting a Polysemantic Neuron.}
To understand how polysemanticity physically manifests, we conduct a detailed case study of Neuron \#652 from the last block of ViT-B. As illustrated in Figure~\ref{fig:polysemantic_neuron_dissection}, we decompose its activations into distinct conceptual clusters. The UMAP projection (Figure~\ref{fig:polysemantic_neuron_dissection}a) and exemplar images (Figure~\ref{fig:polysemantic_neuron_dissection}d) reveal clear semantic heterogeneity, with the neuron responding to entirely unrelated concepts such as snakeskin textures (C1), penguins (C2), and two-wheeled vehicles (C3-C5). The heatmaps (Figure~\ref{fig:polysemantic_neuron_dissection}b, c) further validate that these clusters are structurally well-separated within the model's internal space and possess low semantic overlap, confirming the precise disentanglement by SPICE.

\subsubsection{Formation Mechanisms and Concept Pathways}
\label{sec:attr_path}
\begin{wrapfigure}[13]{r}{0.5\textwidth}
    \centering
    \includegraphics[width=0.9\linewidth]{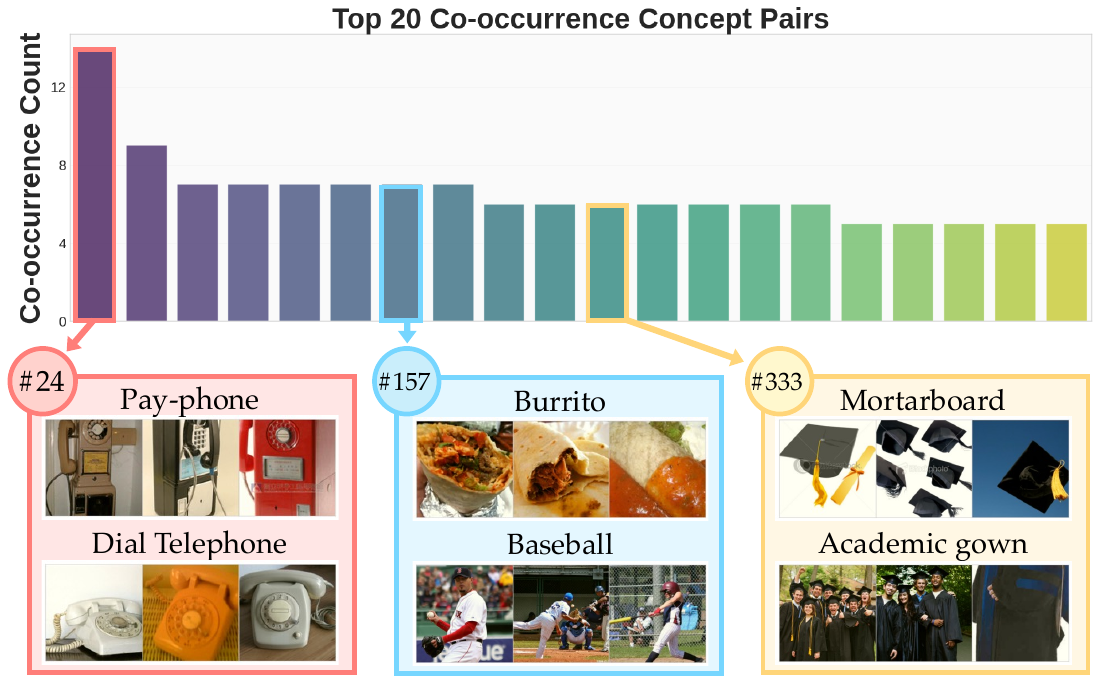}
    \caption{Top 20 co-occurrence concept pairs in ViT-B.}
    \label{fig:co-occur}
\end{wrapfigure}
Having observed that a single neuron can perfectly disentangle into distinct concepts, we next investigate \textit{how} these specific concepts are formed. By measuring label co-occurrences in the final ViT-B block (Figure~\ref{fig:co-occur}), we find that polysemantic neurons arise heterogeneously: some encode highly related concepts (e.g., variants of telephones in neuron \#24), while others combine entirely unrelated categories (e.g., burritos and baseballs in neuron \#157). 

\begin{figure}[t]
    \centering
    \includegraphics[width=0.75\columnwidth]{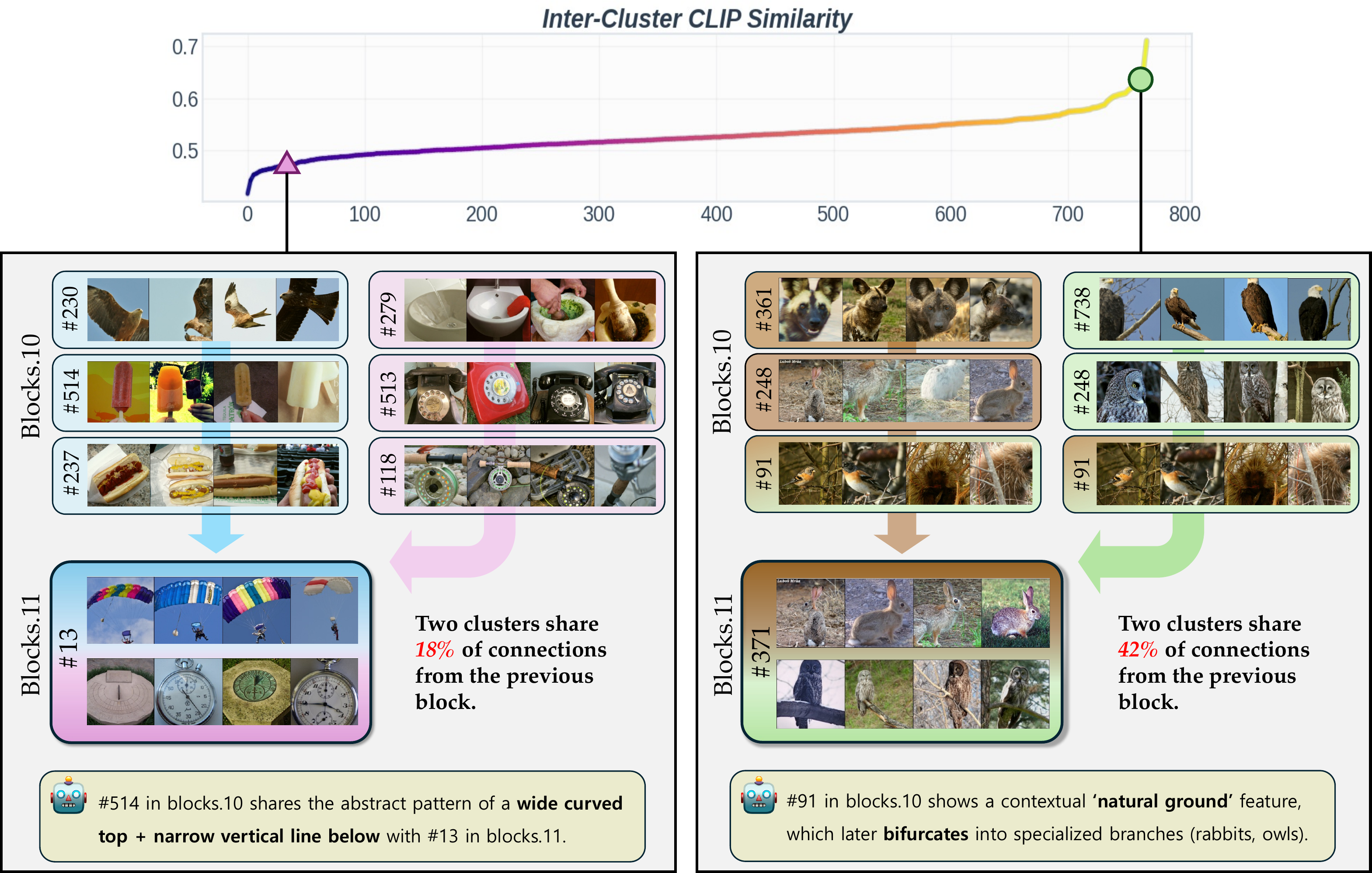}
    \caption{(\textbf{Left}) Inter-cluster CLIP similarity, where the x-axis denotes neurons in ViT-B block 11 sorted by similarity. (\textbf{Middle}, \textbf{Right}) Example of a polysemantic neuron with low inter-cluster similarity, and a high inter-cluster similarity, respectively.}
    \label{fig:attr_path}
\end{figure}

To determine if these distinct forms of polysemanticity stem from different mechanisms, we trace their upstream computational pathways (Figure~\ref{fig:attr_path}). For neuron \#13, which encodes unrelated concepts, we observe largely disjoint upstream connections, with only an 18\% overlap among the top-50 contributing neurons. Conversely, neuron \#371, encoding related concepts, shares 42\% of its attribution pathways, indicating a shared formation circuit. These human-aligned interpretations are further validated by VLM-generated descriptions~\cite{achiam2023gpt, yu2025coe}. Ultimately, these findings reveal that polysemantic neurons are formed through diverse processes—ranging from overlapping semantic circuits to entirely disjoint pathways—which fundamentally dictate their interpretability.

\begin{figure}[!h]
\begin{center}
    \includegraphics[width=0.95\textwidth]{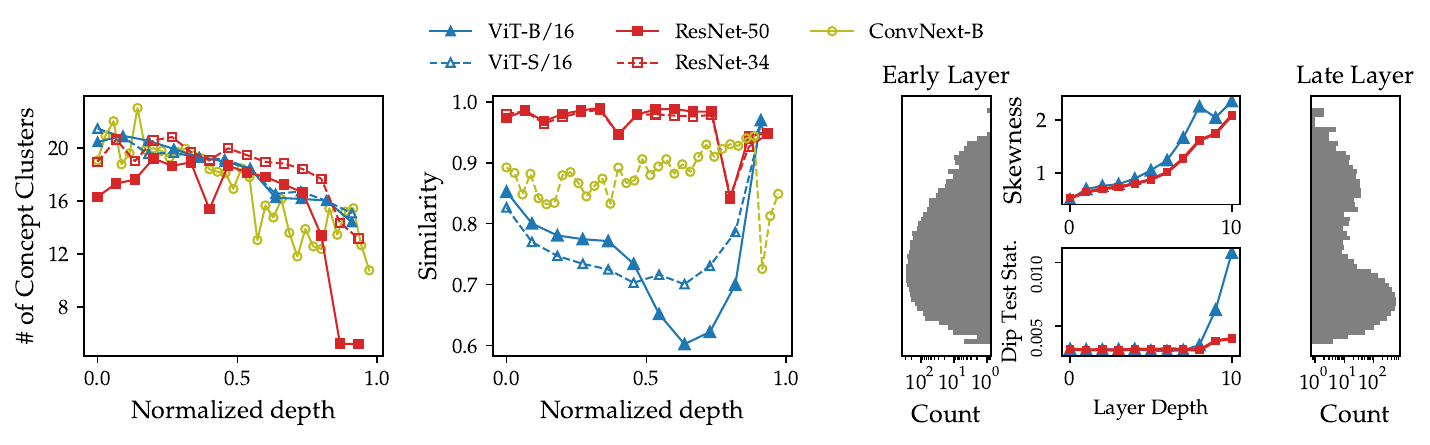}
\end{center}
\caption{Polysemanticity differs across network architectures. \textbf{(Left)} The number of concepts per neuron across layers. \textbf{(Center)} Intra-cluster similarity across layers. \textbf{(Right)} A structural evolution where the similarity distribution shifts from unimodal to bimodal.}
\label{fig:basic_stat}
\end{figure}

\subsubsection{Architectural Inductive Bias.}
Building on these neuron-level mechanisms, we broaden our analysis to examine how macroscopic architectural designs influence polysemantic behavior across the entire network (Figure~\ref{fig:basic_stat}). First, regarding the number of disentangled concepts (left), CNN-based models (ResNet, ConvNeXt) tend to maintain numerous concepts through most of the network, with ResNet-50 in particular showing a pronounced drop in the final blocks, whereas Vision Transformers (ViT) exhibit a more gradual decrease. Second, assessing internal cohesion via intra-cluster similarity (center), CNNs exhibit consistent stability interrupted only by spatial downsampling transitions. Conversely, Transformers display a characteristic U-shaped trend, exploring higher conceptual diversity in middle layers before converging.

These behavioral differences stem from the evolution of feature-similarity distributions (Figure~\ref{fig:basic_stat}, right). While all models begin with unimodal distributions indicating limited separation in early layers, they transition to distinctly bimodal distributions later, reflecting clear concept splits. This multimodality shift, quantified via Hartigan’s Dip Test is most pronounced in ViTs. 
Ultimately, these contrasting trajectories highlight distinct inductive biases: CNNs preserve cohesive concepts driven by local processing, while ViTs undergo mid-layer diversification and late-stage reconsolidation via global self-attention.

\subsubsection{Ablation Studies}
The validity of the aforementioned scientific insights hinges on the robustness of the SPICE framework. Here, we thoroughly evaluate our core design choices.

\begin{table}[t]
    \centering
    \setlength{\tabcolsep}{1.2pt} 
    \renewcommand{\arraystretch}{0.9} 
    \setlength{\belowcaptionskip}{3pt}
    \caption{Separability of ViT-B/16 across activation ranges (Top/Mid/Bottom), showing the same depth-wise trend as Table~\ref{tab:intra_inter_clustering_merged}.}
    \label{tab:activation_range}
    \resizebox{\linewidth}{!}{
    \begin{tabular}{l|ccc|ccc|ccc|ccc|ccc}
        \toprule
        \multirow{2}{*}{\textbf{Method}} & \multicolumn{3}{c|}{\textbf{B2}} & \multicolumn{3}{c|}{\textbf{B4}} & \multicolumn{3}{c|}{\textbf{B6}} & \multicolumn{3}{c|}{\textbf{B8}} & \multicolumn{3}{c}{\textbf{B10}} \\
         & T & M & B & T & M & B & T & M & B & T & M & B & T & M & B \\
        \midrule
        PURE & 1.02 & 1.00 & 1.02 & 1.02 & 1.00 & 1.02 & 1.03 & 1.00 & 1.01 & 1.03 & 1.00 & 1.01 & 1.07 & 1.00 & 1.01 \\
        CPE  & 0.99 & 0.99 & 0.99 & 1.01   & 0.99 & 0.99 & 0.99 & 0.99 & 1.00 & 1.01 & 0.99 & 0.99 & 1.01 & 0.99 & 0.99 \\
        LE   & \textbf{1.20} & \textbf{1.11} & \underline{1.12} & \underline{1.15} & \textbf{1.15} & \underline{1.13} & \underline{1.19} & \textbf{1.18} & \underline{1.13} & \textbf{1.21} & \underline{1.14} & \underline{1.13} & \underline{1.33} & \underline{1.13} & \underline{1.04} \\
        \rowcolor{gray!15} \textbf{Ours} & \underline{1.14} & \underline{1.09} & \textbf{1.12} & \textbf{1.19} & \underline{1.14} & \textbf{1.16} & \textbf{1.24} & \underline{1.18} & \textbf{1.21} & \underline{1.20} & \textbf{1.31} & \textbf{1.29} & \textbf{1.48} & \textbf{1.27} & \textbf{1.41} \\
        \bottomrule
    \end{tabular}
}
\vspace{-1.5em}
\end{table}

\paragraph{Robustness across Activation Ranges.} Standard interpretability evaluation protocols often restrict their analysis to the highly activated range (e.g., Top-10\%) of a neuron. To ensure our method's robustness and comprehensively address the evaluation scope, we expand our quantitative analysis to cover the \textbf{Top}, \textbf{Middle}, and \textbf{Bottom} activation ranges (100 samples each) across varying depths. As shown in Table~\ref{tab:activation_range}, SPICE consistently maintains high separability scores significantly above the PURE and CPE baselines across all depths and ranges. Notably, while LE achieves competitive scores in early layers (e.g., B2), its performance degrades substantially in deeper layers (B10) and lower activation ranges (Bottom: 1.04). In contrast, SPICE demonstrates superior robustness, achieving a dominant score of 1.41 even in the B10 Bottom range. This confirms that SPICE effectively disentangles concepts regardless of activation magnitude, successfully operating even in highly polysemantic and less activated spaces where existing text-prior-based methods fail.

\begin{figure}[!h]
    \centering
    \includegraphics[width=0.7\linewidth,trim={1cm 0cm 0cm 0cm},clip]{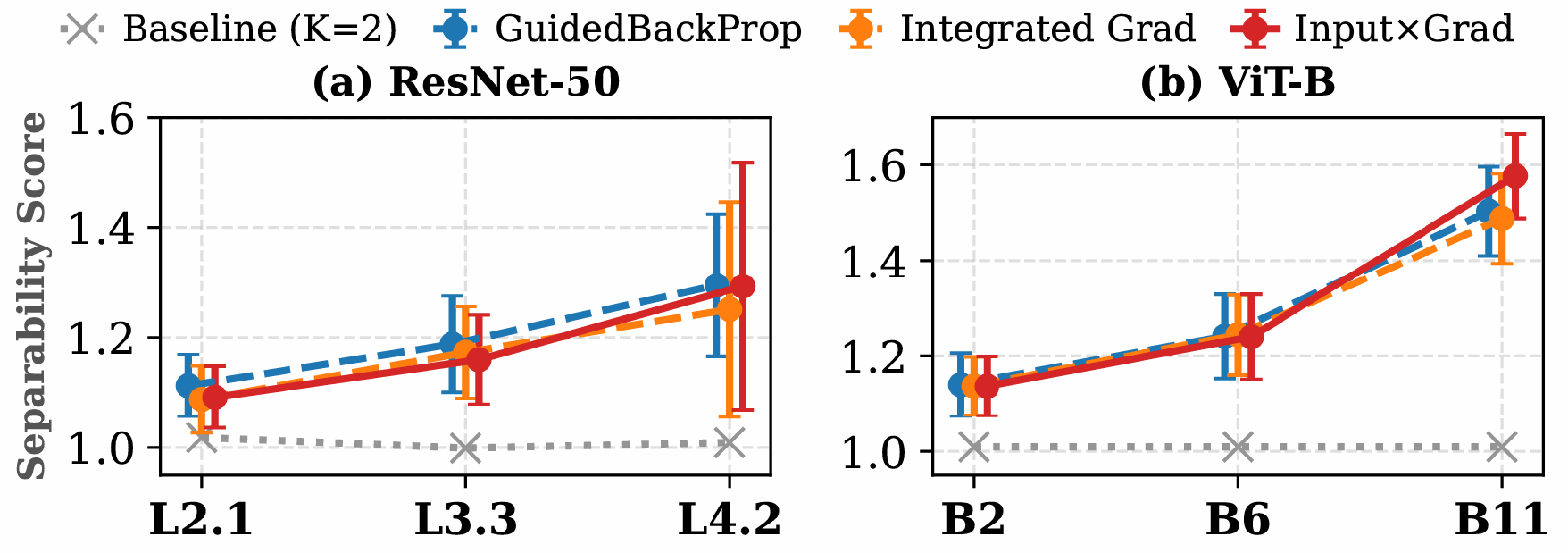}
    \caption{\textbf{Robustness to attribution methods.} SPICE consistently outperforms PURE across Integrated Gradients, GradientShap, Guided Backpropagation, and Input $\times$ Gradient.}
    \label{fig:ablations}
\end{figure}

\begin{table}[!h]
\centering
\small
\setlength{\tabcolsep}{4pt}
\caption{\textbf{Adaptive vs.\ static thresholding.} For each variant we report the separability score and the coverage (\# of 100 randomly sampled neurons yielding at least one cohesive cluster). \emph{Fixed $K\!=\!2$} accepts every cluster (no cohesion threshold $\Rightarrow$ 100\% coverage by construction). \emph{Static $\tau$} applies a global cosine-cohesion threshold; high $\tau$ inflates separability but drops coverage to near zero because most neurons cannot meet the threshold. \emph{Adaptive (SPICE)} sets $\tau$ per neuron, achieving near-100\% coverage with strong separability.}
\label{tab:adaptive_vs_static}
\begin{tabular}{l|cc|cc}
\toprule
& \multicolumn{2}{c|}{\textbf{ResNet-50 (L3.3)}} & \multicolumn{2}{c}{\textbf{ViT-B (B6)}} \\
\textbf{Variant} & Score & Coverage & Score & Coverage \\
\midrule
Fixed $K\!=\!2$ (no $\tau$)            & 1.00 & 100/100 & 1.03 & 100/100 \\
\midrule
Static $\tau \geq 0.9$                  & 1.56 & 6/100   & 2.18 & 7/100   \\
Static $\tau \geq 0.8$                  & 1.56 & 9/100   & 2.18 & 10/100  \\
Static $\tau \geq 0.7$                  & 1.39 & 27/100  & 1.88 & 21/100  \\
\midrule
\rowcolor{gray!15}
\textbf{Adaptive $\tau$ (SPICE)}        & \textbf{1.15} & \textbf{100/100} & \textbf{1.25} & \textbf{100/100} \\
\bottomrule
\end{tabular}
\end{table}

\paragraph{Robustness to Attribution Methods.} To validate our core design choices, we first examine whether SPICE's disentanglement relies on the specific attribution method used to compute the representational footprints. As shown in Figure~\ref{fig:ablations}, SPICE consistently outperforms the PURE baseline regardless of the attribution method employed---including Integrated Gradients~\cite{IntegratedGradient}, GradientShap~\cite{GradientShap}, Guided Backpropagation~\cite{GuidedBackProb}, and Input $\times$ Gradient. This confirms that the significant performance gain stems from our framework itself, rather than an artifact of a specific metric. We adopt Input $\times$ Gradient as our default choice due to its efficiency.

\paragraph{Adaptive vs. Static Thresholding.} Finally, we evaluate the necessity of our adaptive clustering algorithm by comparing it against two natural alternatives: (i) accepting every cluster from a fixed $K\!=\!2$ partition (no cohesion threshold), and (ii) applying a single \emph{global} cohesion threshold $\tau$ shared across all neurons. We measure both separability and \emph{coverage}, defined as the number of neurons (out of 100 randomly sampled) for which at least one cluster passes the cohesion test; fixed-$K$ trivially achieves 100\% coverage by construction since it imposes no threshold. As shown in Table~\ref{tab:adaptive_vs_static}, a strict global $\tau$ (e.g., $\tau \geq 0.9$) yields high raw separability but applies to only $6$--$7$ out of $100$ neurons, since after attribution normalization the pairwise similarities concentrate in a narrow range well below $0.9$ and most neurons have no cluster cohesive enough to clear the threshold. Loosening $\tau$ raises coverage but degrades the score. SPICE's per-neuron adaptive $\tau$ instead achieves near-100\% coverage while maintaining strong separability over the same population, demonstrating that the threshold must be calibrated per neuron to generalize interpretation across the entire network. 

\section{Limitations}
While SPICE generally yields superior separability, the utility of more advanced attribution methods warrants further investigation. Moreover, the lower separability observed in the early layers (e.g., B2) of ViT-B highlights a structural constraint of the CLIP verification protocol itself as early layers capture low-level features that are not easily distinguishable within CLIP's high-level semantic space. This underscores that numerical assessments of concept quality heavily depend on the external semantic space used for verification. Despite its generality, our method presents practical constraints: generating human-understandable labels relies on external VLMs, and the iterative clustering process is more computationally intensive than single-pass methods. These challenges motivate clear avenues for future work, such as developing strategies to harness VLM knowledge while compensating for representational misalignment. Ultimately, the framework can be extended toward circuit-level disentanglement, moving beyond single neurons to systematically trace the formation mechanisms of concepts along their pathways.

\section{Conclusion}
In this work, we addressed the challenge of polysemanticity—the tendency of a single neuron to encode multiple, often unrelated concepts—which obscures the functional interpretation in deep models. We introduce SPICE (Simple Polysemantic Feature Interpretation via Clustering-based Explanation), a simple and generalizable framework that overcomes the architectural constraints and heuristic assumptions (e.g., fixed K) of prior approaches. SPICE effectively disentangles the visual concepts encoded within individual neurons, enabling the first systematic comparison of polysemanticity across both CNNs and Transformers. By automatically determining the number of concept clusters per neuron, it eliminates reliance on manual hyperparameters and provides a scalable analysis pipeline suitable for large modern architectures. Using SPICE, we conduct a comprehensive analysis across diverse deep vision architectures. We believe this work highlights the importance of shifting interpretability from single-neuron analyses to the concept clusters they encode, providing a more faithful lens through which to understand the internal mechanisms of deep visual models.


\section*{Acknowledgements} This work was supported by the Institute for Information \& Communications Technology Planning \& Evaluation (IITP) grant funded by the Korea government (MSIT) (RS-2019-II190075, Artificial Intelligence Graduate School Support Program (KAIST); RS-2024-00457882, AI Research Hub Project; and RS-2022-II220984, Development of Artificial Intelligence Technology for Personalized Plug-and-Play Explanation and Verification of Explanation).

%
%
\bibliographystyle{splncs04}
\bibliography{main}

@String(PAMI  = {IEEE Trans. Pattern Anal. Mach. Intell.})

@String(CVPR  = {IEEE Conf. Comput. Vis. Pattern Recog.})

@String(ICCV  = {Int. Conf. Comput. Vis.})

@String(ECCV  = {Eur. Conf. Comput. Vis.})

@String(ICML  = {Int. Conf. Mach. Learn.})

@String(ICLR  = {Int. Conf. Learn. Represent.})

@String(PAMI  = {IEEE TPAMI})

@String(CVPR  = {CVPR})

@String(ICCV  = {ICCV})

@String(ECCV  = {ECCV})

@String(ICML  = {ICML})

@String(ICLR  = {ICLR})

@inproceedings{bau2017network,
  title={Network dissection: Quantifying interpretability of deep visual representations},
  author={Bau, David and Zhou, Bolei and Khosla, Aditya and Oliva, Aude and Torralba, Antonio},
  booktitle={Proceedings of the IEEE conference on computer vision and pattern recognition},
  pages={6541--6549},
  year={2017}
}

@article{elhage2022toy,
  title={Toy models of superposition},
  author={Elhage, Nelson and Hume, Tristan and Olsson, Catherine and Schiefer, Nicholas and Henighan, Tom and Kravec, Shauna and Hatfield-Dodds, Zac and Lasenby, Robert and Drain, Dawn and Chen, Carol and others},
  journal={arXiv preprint arXiv:2209.10652},
  year={2022}
}

@inproceedings{dreyer2024pure,
  title={Pure: Turning polysemantic neurons into pure features by identifying relevant circuits},
  author={Dreyer, Maximilian and Purelku, Erblina and Vielhaben, Johanna and Samek, Wojciech and Lapuschkin, Sebastian},
  booktitle={Proceedings of the IEEE/CVF Conference on Computer Vision and Pattern Recognition},
  pages={8212--8217},
  year={2024}
}

@article{cunningham2023sparse,
  title={Sparse autoencoders find highly interpretable features in language models},
  author={Cunningham, Hoagy and Ewart, Aidan and Riggs, Logan and Huben, Robert and Sharkey, Lee},
  journal={arXiv preprint arXiv:2309.08600},
  year={2023}
}

@inproceedings{thasarathan2025universal,
  title={Universal sparse autoencoders: Interpretable cross-model concept alignment},
  author={Thasarathan, Harrish and Forsyth, Julian and Fel, Thomas and Kowal, Matthew and Derpanis, Konstantinos G},
  booktitle={Forty-second International Conference on Machine Learning},
  year={2025}
}

@inproceedings{OikarinenW23CLIPdissect,
  author       = {Tuomas P. Oikarinen and
                  Tsui{-}Wei Weng},
  title        = {CLIP-Dissect: Automatic Description of Neuron Representations in Deep
                  Vision Networks},
  booktitle    = {The Eleventh International Conference on Learning Representations,
                  {ICLR} 2023, Kigali, Rwanda, May 1-5, 2023},
  publisher    = {OpenReview.net},
  year         = {2023},
  url          = {https://openreview.net/forum?id=iPWiwWHc1V},
  bibsource    = {dblp computer science bibliography, https://dblp.org}
}

@article{arrieta2020explainable,
  title={Explainable Artificial Intelligence (XAI): Concepts, taxonomies, opportunities and challenges toward responsible AI},
  author={Arrieta, Alejandro Barredo and D{\'\i}az-Rodr{\'\i}guez, Natalia and Del Ser, Javier and Bennetot, Adrien and Tabik, Siham and Barbado, Alberto and Garc{\'\i}a, Salvador and Gil-L{\'o}pez, Sergio and Molina, Daniel and Benjamins, Richard and others},
  journal={Information fusion},
  volume={58},
  pages={82--115},
  year={2020},
  publisher={Elsevier}
}

@article{olah2017feature,
  title={Feature visualization},
  author={Olah, Chris and Mordvintsev, Alexander and Schubert, Ludwig},
  journal={Distill},
  volume={2},
  number={11},
  pages={e7},
  year={2017}
}

@article{olah2020zoom,
  title={Zoom in: An introduction to circuits},
  author={Olah, Chris and Cammarata, Nick and Schubert, Ludwig and Goh, Gabriel and Petrov, Michael and Carter, Shan},
  journal={Distill},
  volume={5},
  number={3},
  pages={e00024--001},
  year={2020}
}

@article{zhou2017places,
  title={Places: A 10 million image database for scene recognition},
  author={Zhou, Bolei and Lapedriza, Agata and Khosla, Aditya and Oliva, Aude and Torralba, Antonio},
  journal=PAMI,
  volume={40},
  number={6},
  pages={1452--1464},
  year={2017}
}

@article{thomee2016yfcc100m,
  title={{YFCC100M}: The new data in multimedia research},
  author={Thomee, Bart and Shamma, David A and Friedland, Gerald and Elizalde, Benjamin and Ni, Karl and Poland, Douglas and Borth, Damian and Li, Li-Jia},
  journal={Communications of the ACM},
  volume={59},
  number={2},
  pages={64--73},
  year={2016}
}

@inproceedings{bousselham2024legrad,
  title={{LeGrad}: An explainability method for vision transformers via feature formation sensitivity},
  author={Bousselham, Walid and Boggust, Angie and Chaybouti, Sofian and Strobelt, Hendrik and Kuehne, Hilde},
  booktitle=ICCV,
  pages={20336--20345},
  year={2025}
}

@article{wrobel2024omenn,
  title={{OMENN}: One Matrix to Explain Neural Networks},
  author={Wr{\'o}bel, Adam and Janusz, Miko{\l}aj and Zieli{\'n}ski, Bartosz and Rymarczyk, Dawid},
  journal={arXiv preprint arXiv:2412.02399},
  year={2024}
}

@inproceedings{zeiler2014visualizing,
  title={Visualizing and understanding convolutional networks},
  author={Zeiler, Matthew D and Fergus, Rob},
  booktitle=ECCV,
  pages={818--833},
  year={2014},
  organization={Springer}
}

@article{lecun2015deep,
  title={Deep learning},
  author={LeCun, Yann and Bengio, Yoshua and Hinton, Geoffrey},
  journal={Nature},
  volume={521},
  number={7553},
  pages={436--444},
  year={2015}
}

@article{fel2025archetypal,
  title={Archetypal {SAE}: Adaptive and stable dictionary learning for concept extraction in large vision models},
  author={Fel, Thomas and Lubana, Ekdeep Singh and Prince, Jacob S and Kowal, Matthew and Boutin, Victor and Papadimitriou, Isabel and Wang, Binxu and Wattenberg, Martin and Ba, Demba and Konkle, Talia},
  journal={arXiv preprint arXiv:2502.12892},
  year={2025}
}

@article{bereska2024mechanistic,
  title={Mechanistic interpretability for AI safety--a review},
  author={Bereska, Leonard and Gavves, Efstratios},
  journal={arXiv preprint arXiv:2404.14082},
  year={2024}
}

@article{elhage2021mathematical,
  title={A mathematical framework for transformer circuits},
  author={Elhage, Nelson and Nanda, Neel and Olsson, Catherine and Henighan, Tom and Joseph, Nicholas and Mann, Ben and Askell, Amanda and Bai, Yuntao and Chen, Anna and Conerly, Tom and others},
  journal={Transformer Circuits Thread},
  volume={1},
  number={1},
  pages={12},
  year={2021}
}

@inproceedings{kalibhat2023identifying,
  title={Identifying interpretable subspaces in image representations},
  author={Kalibhat, Neha and Bhardwaj, Shweta and Bruss, C Bayan and Firooz, Hamed and Sanjabi, Maziar and Feizi, Soheil},
  booktitle={International Conference on Machine Learning},
  pages={15623--15638},
  year={2023},
  organization={PMLR}
}

@inproceedings{NeurFlow,
  author       = {Tue Minh Cao and
                  Nhat Hoang{-}Xuan and
                  Hieu H. Pham and
                  Phi Le Nguyen and
                  My T. Thai},
  title        = {NeurFlow: Interpreting Neural Networks through Neuron Groups and Functional
                  Interactions},
  booktitle    = {The Thirteenth International Conference on Learning Representations,
                  {ICLR} 2025, Singapore, April 24-28, 2025},
  publisher    = {OpenReview.net},
  year         = {2025},
  url          = {https://openreview.net/forum?id=GdbQyFOUlJ},
  bibsource    = {dblp computer science bibliography, https://dblp.org}
}

@article{wang2022interpretability,
  title={Interpretability in the wild: a circuit for indirect object identification in gpt-2 small},
  author={Wang, Kevin and Variengien, Alexandre and Conmy, Arthur and Shlegeris, Buck and Steinhardt, Jacob},
  journal={arXiv preprint arXiv:2211.00593},
  year={2022}
}

@inproceedings{VCC,
  title={Visual concept connectome (vcc): Open world concept discovery and their interlayer connections in deep models},
  author={Kowal, Matthew and Wildes, Richard P and Derpanis, Konstantinos G},
  booktitle={Proceedings of the IEEE/CVF Conference on Computer Vision and Pattern Recognition},
  pages={10895--10905},
  year={2024}
}

@article{marshall2024understanding,
  title={Understanding polysemanticity in neural networks through coding theory},
  author={Marshall, Simon C and Kirchner, Jan H},
  journal={arXiv preprint arXiv:2401.17975},
  year={2024}
}

@article{achiam2023gpt,
  title={Gpt-4 technical report},
  author={Achiam, Josh and Adler, Steven and Agarwal, Sandhini and Ahmad, Lama and Akkaya, Ilge and Aleman, Florencia Leoni and Almeida, Diogo and Altenschmidt, Janko and Altman, Sam and Anadkat, Shyamal and others},
  journal={arXiv preprint arXiv:2303.08774},
  year={2023}
}

@misc{zhai2023SigLIP,
      title={Sigmoid Loss for Language Image Pre-Training}, 
      author={Xiaohua Zhai and Basil Mustafa and Alexander Kolesnikov and Lucas Beyer},
      year={2023},
      eprint={2303.15343},
      archivePrefix={arXiv},
      primaryClass={cs.CV},
      url={https://arxiv.org/abs/2303.15343}, 
}

@inproceedings{oikarinen2024linear,
        title={Linear Explanations for Individual Neurons},
        author={Oikarinen, Tuomas and Weng, Tsui-Wei},
        booktitle={International Conference on Machine Learning},
        year={2024}
      }

@inproceedings{yu2025coe,
  title={CoE: Chain-of-Explanation via Automatic Visual Concept Circuit Description and Polysemanticity Quantification},
  author={Yu, Wenlong and Wang, Qilong and Liu, Chuang and Li, Dong and Hu, Qinghua},
  booktitle={Proceedings of the Computer Vision and Pattern Recognition Conference},
  pages={4364--4374},
  year={2025}
}

@inproceedings{hesse2025disentangling,
  title={Disentangling Polysemantic Channels in Convolutional Neural Networks},
  author={Hesse, Robin and Fischer, Jonas and Schaub-Meyer, Simone and Roth, Stefan},
  booktitle={Proceedings of the Computer Vision and Pattern Recognition Conference},
  pages={4799--4803},
  year={2025}
}

@inproceedings{shrikumar2017learning,
  title={Learning important features through propagating activation differences},
  author={Shrikumar, Avanti and Greenside, Peyton and Kundaje, Anshul},
  booktitle={International conference on machine learning},
  pages={3145--3153},
  year={2017},
  organization={PMlR}
}

@inproceedings{selvaraju2017grad,
  title={Grad-cam: Visual explanations from deep networks via gradient-based localization},
  author={Selvaraju, Ramprasaath R and Cogswell, Michael and Das, Abhishek and Vedantam, Ramakrishna and Parikh, Devi and Batra, Dhruv},
  booktitle={Proceedings of the IEEE international conference on computer vision},
  pages={618--626},
  year={2017}
}

@article{simonyan2014visualising,
  title={Visualising image classification models and saliency maps},
  author={Simonyan, Karen and Vedaldi, Andrea and Zisserman, Andrew},
  journal={Deep Inside Convolutional Networks},
  volume={2},
  number={2},
  year={2014}
}

@article{godey2024anisotropy,
  title={Anisotropy is inherent to self-attention in transformers},
  author={Godey, Nathan and de la Clergerie, {\'E}ric and Sagot, Beno{\^\i}t},
  journal={arXiv preprint arXiv:2401.12143},
  year={2024}
}

@inproceedings{razzhigaev2024shape,
  title={The shape of learning: Anisotropy and intrinsic dimensions in transformer-based models},
  author={Razzhigaev, Anton and Mikhalchuk, Matvey and Goncharova, Elizaveta and Oseledets, Ivan and Dimitrov, Denis and Kuznetsov, Andrey},
  booktitle={Findings of the Association for Computational Linguistics: EACL 2024},
  pages={868--874},
  year={2024}
}

@article{bansal2018can,
  title={Can we gain more from orthogonality regularizations in training deep networks?},
  author={Bansal, Nitin and Chen, Xiaohan and Wang, Zhangyang},
  journal={Advances in Neural Information Processing Systems},
  volume={31},
  year={2018}
}

@article{elhage2023privileged,
  title={Privileged bases in the transformer residual stream},
  author={Elhage, Nelson and Lasenby, Robert and Olah, Christopher},
  journal={Transformer Circuits Thread},
  pages={24},
  year={2023}
}

@inproceedings{
kopf2024cosy,
title={CoSy: Evaluating Textual Explanations of Neurons},
author={Laura Kopf and Philine Lou Bommer and Anna Hedstr{\"o}m and Sebastian Lapuschkin and Marina MC H{\"o}hne and Kirill Bykov},
booktitle={The Thirty-eighth Annual Conference on Neural Information Processing Systems},
year={2024},
url={https://openreview.net/forum?id=R0bnWrpIeN}
}

@article{hartigan1985dip,
  title={The dip test of unimodality},
  author={Hartigan, John A and Hartigan, Pamela M},
  journal={The annals of Statistics},
  pages={70--84},
  year={1985},
  publisher={JSTOR}
}

@inproceedings{
dosovitskiy2021an,
title={An Image is Worth 16x16 Words: Transformers for Image Recognition at Scale},
author={Alexey Dosovitskiy and Lucas Beyer and Alexander Kolesnikov and Dirk Weissenborn and Xiaohua Zhai and Thomas Unterthiner and Mostafa Dehghani and Matthias Minderer and Georg Heigold and Sylvain Gelly and Jakob Uszkoreit and Neil Houlsby},
booktitle={International Conference on Learning Representations},
year={2021},
url={https://openreview.net/forum?id=YicbFdNTTy}
}

@article{He2015DeepRL,
  title={Deep Residual Learning for Image Recognition},
  author={Kaiming He and X. Zhang and Shaoqing Ren and Jian Sun},
  journal={2016 IEEE Conference on Computer Vision and Pattern Recognition (CVPR)},
  year={2015},
  pages={770-778},
  url={https://api.semanticscholar.org/CorpusID:206594692}
}

@inproceedings{liu2022convnet,
  title={A convnet for the 2020s},
  author={Liu, Zhuang and Mao, Hanzi and Wu, Chao-Yuan and Feichtenhofer, Christoph and Darrell, Trevor and Xie, Saining},
  booktitle={Proceedings of the IEEE/CVF conference on computer vision and pattern recognition},
  pages={11976--11986},
  year={2022}
}

@inproceedings{imagenet_cvpr09,
        AUTHOR = {Deng, J. and Dong, W. and Socher, R. and Li, L.-J. and Li, K. and Fei-Fei, L.},
        TITLE = {{ImageNet: A Large-Scale Hierarchical Image Database}},
        BOOKTITLE = {CVPR09},
        YEAR = {2009},
        BIBSOURCE = "http://www.image-net.org/papers/imagenet_cvpr09.bib"
}

@article{achtibat2023attribution,
  title={From attribution maps to human-understandable explanations through concept relevance propagation},
  author={Achtibat, Reduan and Dreyer, Maximilian and Eisenbraun, Ilona and Bosse, Sebastian and Wiegand, Thomas and Samek, Wojciech and Lapuschkin, Sebastian},
  journal={Nature Machine Intelligence},
  volume={5},
  number={9},
  pages={1006--1019},
  year={2023},
  publisher={Nature Publishing Group UK London}
}

@article{simeoni2025dinov3,
  title={Dinov3},
  author={Sim{\'e}oni, Oriane and Vo, Huy V and Seitzer, Maximilian and Baldassarre, Federico and Oquab, Maxime and Jose, Cijo and Khalidov, Vasil and Szafraniec, Marc and Yi, Seungeun and Ramamonjisoa, Micha{\"e}l and others},
  journal={arXiv preprint arXiv:2508.10104},
  year={2025}
}

@inproceedings{bossard2014food,
  title={Food-101--mining discriminative components with random forests},
  author={Bossard, Lukas and Guillaumin, Matthieu and Van Gool, Luc},
  booktitle={European conference on computer vision},
  pages={446--461},
  year={2014},
  organization={Springer}
}

@article{ball1965isodata,
  title={ISODATA, a novel method of data analysis and pattern classification},
  author={Ball, Geoffrey H and Hall, David J},
  year={1965}
}

@inproceedings{campello2013density,
  title={Density-based clustering based on hierarchical density estimates},
  author={Campello, Ricardo JGB and Moulavi, Davoud and Sander, J{\"o}rg},
  booktitle={Pacific-Asia conference on knowledge discovery and data mining},
  pages={160--172},
  year={2013},
  organization={Springer}
}

@inproceedings{GradientShap,
  author       = {Scott M. Lundberg and
                  Su{-}In Lee},
  editor       = {Isabelle Guyon and
                  Ulrike von Luxburg and
                  Samy Bengio and
                  Hanna M. Wallach and
                  Rob Fergus and
                  S. V. N. Vishwanathan and
                  Roman Garnett},
  title        = {A Unified Approach to Interpreting Model Predictions},
  booktitle    = {Advances in Neural Information Processing Systems 30: Annual Conference
                  on Neural Information Processing Systems 2017, December 4-9, 2017,
                  Long Beach, CA, {USA}},
  pages        = {4765--4774},
  year         = {2017},
  url          = {https://proceedings.neurips.cc/paper/2017/hash/8a20a8621978632d76c43dfd28b67767-Abstract.html},
  bibsource    = {dblp computer science bibliography, https://dblp.org}
}

@inproceedings{GuidedBackProb,
  author       = {Jost Tobias Springenberg and
                  Alexey Dosovitskiy and
                  Thomas Brox and
                  Martin A. Riedmiller},
  editor       = {Yoshua Bengio and
                  Yann LeCun},
  title        = {Striving for Simplicity: The All Convolutional Net},
  booktitle    = {3rd International Conference on Learning Representations, {ICLR} 2015,
                  San Diego, CA, USA, May 7-9, 2015, Workshop Track Proceedings},
  year         = {2015},
  url          = {http://arxiv.org/abs/1412.6806},
  bibsource    = {dblp computer science bibliography, https://dblp.org}
}

@inproceedings{IntegratedGradient,
  author       = {Mukund Sundararajan and
                  Ankur Taly and
                  Qiqi Yan},
  editor       = {Doina Precup and
                  Yee Whye Teh},
  title        = {Axiomatic Attribution for Deep Networks},
  booktitle    = {Proceedings of the 34th International Conference on Machine Learning,
                  {ICML} 2017, Sydney, NSW, Australia, 6-11 August 2017},
  series       = {Proceedings of Machine Learning Research},
  volume       = {70},
  pages        = {3319--3328},
  publisher    = {{PMLR}},
  year         = {2017},
  url          = {http://proceedings.mlr.press/v70/sundararajan17a.html},
  bibsource    = {dblp computer science bibliography, https://dblp.org}
}

@inproceedings{radford2021learning,
  author       = {Alec Radford and
                  Jong Wook Kim and
                  Chris Hallacy and
                  Aditya Ramesh and
                  Gabriel Goh and
                  Sandhini Agarwal and
                  Girish Sastry and
                  Amanda Askell and
                  Pamela Mishkin and
                  Jack Clark and
                  Gretchen Krueger and
                  Ilya Sutskever},
  editor       = {Marina Meila and
                  Tong Zhang},
  title        = {Learning Transferable Visual Models From Natural Language Supervision},
  booktitle    = {Proceedings of the 38th International Conference on Machine Learning,
                  {ICML} 2021, 18-24 July 2021, Virtual Event},
  series       = {Proceedings of Machine Learning Research},
  volume       = {139},
  pages        = {8748--8763},
  publisher    = {{PMLR}},
  year         = {2021},
  url          = {http://proceedings.mlr.press/v139/radford21a.html},
  bibsource    = {dblp computer science bibliography, https://dblp.org}
}

@inproceedings{huang2017densely,
  author       = {Gao Huang and
                  Zhuang Liu and
                  Laurens van der Maaten and
                  Kilian Q. Weinberger},
  title        = {Densely Connected Convolutional Networks},
  booktitle    = {2017 {IEEE} Conference on Computer Vision and Pattern Recognition,
                  {CVPR} 2017, Honolulu, HI, USA, July 21-26, 2017},
  pages        = {2261--2269},
  publisher    = {{IEEE} Computer Society},
  year         = {2017},
  url          = {https://doi.org/10.1109/CVPR.2017.243},
  doi          = {10.1109/CVPR.2017.243},
  bibsource    = {dblp computer science bibliography, https://dblp.org}
}

@inproceedings{kwon2025granular,
  title={Granular Concept Circuits: Toward a Fine-Grained Circuit Discovery for Concept Representations},
  author={Kwon, Dahee and Lee, Sehyun and Choi, Jaesik},
  booktitle={Proceedings of the IEEE/CVF International Conference on Computer Vision},
  pages={2356--2365},
  year={2025}
}

\clearpage
\title{SPICE: Simple Polysemantic Feature Interpretation via Clustering-based Explanation\\ Supplementary Material}

\titlerunning{SPICE: Simple Polysemantic Feature Interpretation}

\author{
Sehyun Lee\inst{1}\orcidlink{0009-0002-1711-9211} \and
Dahee Kwon\inst{1}\orcidlink{0009-0002-6395-8059} \and
Damin Lee\inst{1}\orcidlink{0009-0004-9380-1006} \and
Jaesik Choi\inst{1,2}\orcidlink{0000-0002-4663-3263}
}

\authorrunning{S.~Lee et al.}

\institute{
Korea Advanced Institute of Science and Technology (KAIST), Daejeon, Republic of Korea\\
\email{\{sehyun.lee, daheekwon, jaesik.choi\}@kaist.ac.kr, xxdamin@gmail.com}
\and
INEEJI, Seongnam, Republic of Korea
}

\maketitle

\section{Experimental Settings}
\label{appx:sec:settings}
\setcounter{page}{1}

\paragraph{Models}
We apply our framework to widely-used models: ViT~\cite{dosovitskiy2021an}, ResNet~\cite{He2015DeepRL}, ConvNeXT~\cite{liu2022convnet}, DenseNet~\cite{huang2017densely}, and CLIP ViT~\cite{radford2021learning}. All models are pretrained on ImageNet~\cite{imagenet_cvpr09}, except CLIP (pretrained on image-text pairs). We analyzed specific layers across these architectures: all bottlenecks in ResNet-50 residual blocks, various dense blocks in DenseNet, all 12 transformer blocks in ViT-B and CLIP ViT-B's visual encoder, and all blocks across the 4 stages of ConvNeXt.

\paragraph{Baselines} We compare our method against three established baselines: PURE~\cite{dreyer2024pure}, LE~\cite{oikarinen2024linear}, and CPE~\cite{yu2025coe}. For all baselines and our method, we conduct the analysis using the top-100 highly activated samples for each neuron to ensure a consistent evaluation of local semantics.

\textbf{PURE} is a pioneering approach that virtually disentangles polysemantic neurons using attribution methods. While its underlying Concept Relevance Propagation (CRP)~\cite{achtibat2023attribution} was originally designed specifically for CNNs, we adapted its implementation for Transformer models to ensure a comprehensive comparison across diverse architectures. Furthermore, PURE originally requires manual specification of the number of concepts. To ensure comparability, we instead automatically determine the optimal number of concepts ($K$) using the Silhouette Score, searching over the candidate set \{1, 2, 3, 4, 5, 6, 7, 8, 9, 10, 20, 30, 40, 50\}. \textbf{LE} computes sparse CPEfficients via linear approximation over a predefined concept set across the entire activation range. While we adhere to the original protocol by fitting the model over the full activation spectrum, we restrict the final evaluation to the concepts associated with the highly activated range to ensure a fair comparison. Since LE allows for the assignment of multiple concepts to a single sample, we allocated concepts accordingly for each sample to facilitate direct comparison. We note that LE's variance in Table~\ref{tab:intra_inter_clustering_merged} is markedly higher than the other baselines; this is a direct consequence of fitting a linear approximation on the narrow activation ranges characteristic of highly polysemantic neurons, where small perturbations in the fit translate into large swings in the per-neuron score. We deliberately retain this regime to match the original LE protocol for a fair comparison, and additionally report the middle/bottom activation ranges in Table~\ref{tab:activation_range} of the main paper for completeness. \textbf{CPE}~\cite{yu2025coe} is evaluated following its original Chain-of-Explanation framework. We use the acronym CPE (Concept Polysemanticity Entropy) in the main text to emphasize its core metric. Concept extraction is performed using gpt-4~\cite{achiam2023gpt}.

\section{Ablation Studies and Sensitivity Analysis}
\label{appx:sec:ablation}
This section evaluates the impact of individual components within our framework, including advanced attribution techniques, threshold, normalization, and different clustering methods.

\begin{figure}[!b]
    \centering
    \begin{subfigure}{0.9\linewidth}
        \centering
        \includegraphics[width=\linewidth]{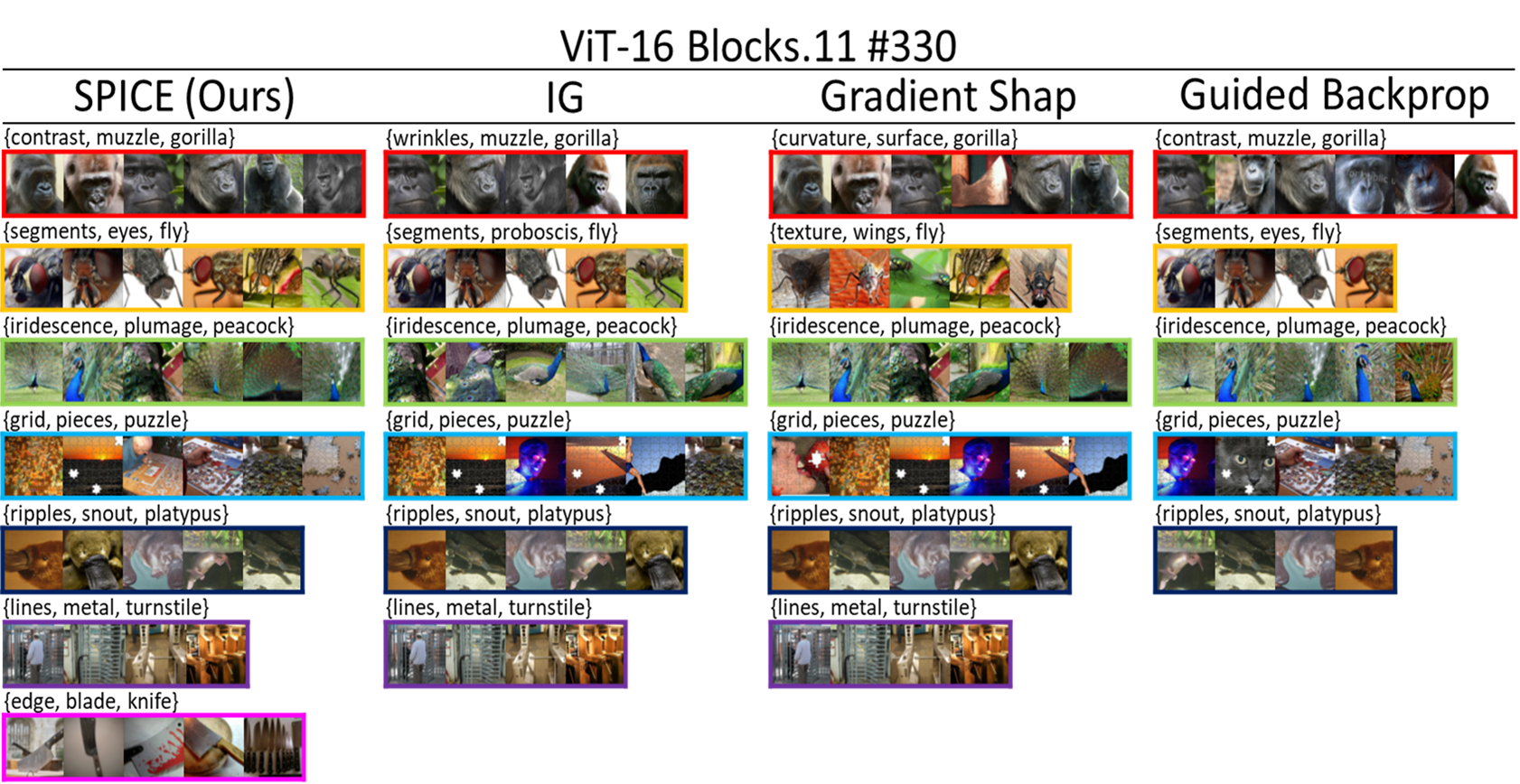}
        \caption{Neuron \#330 in Block 11 of ViT-B/16 under different attribution methods.}
        \label{appx:fig:attr_vit}
    \end{subfigure}
        
    \begin{subfigure}{0.9\linewidth}
        \centering
        \includegraphics[width=\linewidth]{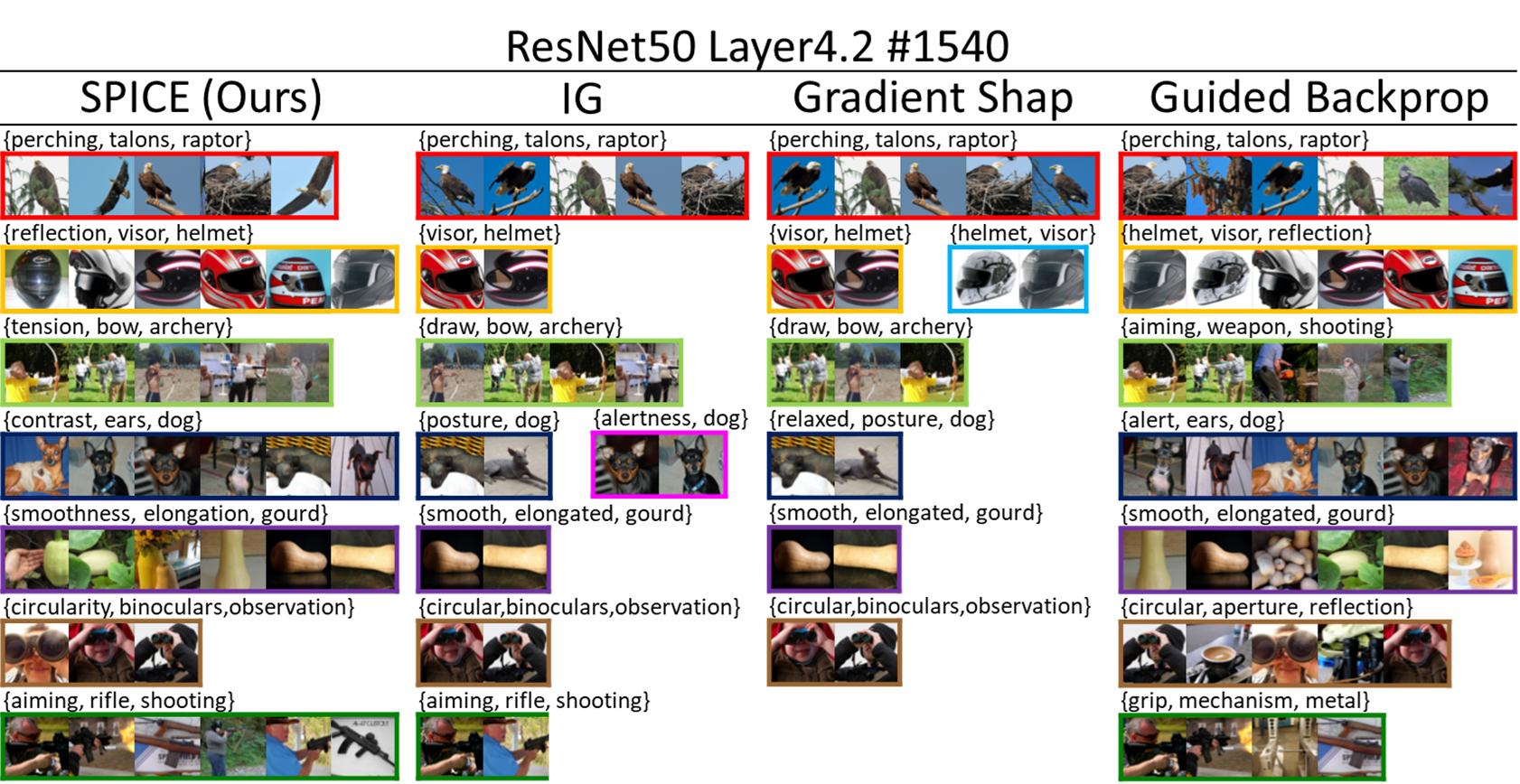}
        \caption{Neuron \#1540 in Layer 4.2 of ResNet-50 under different attribution methods.}
        \label{appx:fig:attr_resnet}
    \end{subfigure}
    \caption{Comparison of concept clusters under different attribution methods for ViT-B/16 and ResNet-50.}
    \label{appx:fig:attr_combined}
\end{figure}

\subsection{Robustness to Attribution Methods}
\label{appx:subsec:robustness_attr}
To qualitatively complement the quantitative robustness results presented in the main text, we provide visual comparisons of the concept clusters discovered using different gradient-based attribution methods, including Integrated Gradients~\cite{IntegratedGradient}, GradientShap~\cite{GradientShap}, and Guided Backprop~\cite{GuidedBackProb}.

As illustrated in Figure~\ref{appx:fig:attr_vit} (ViT-B/16) and Figure~\ref{appx:fig:attr_resnet} (ResNet-50), our proposed approach (using Input $\times$ Gradient) consistently yields visually coherent groupings. For instance, in the ViT-B/16 example, GradientShap tends to produce semantically mixed clusters, whereas SPICE with Input $\times$ Gradient maintains highly cohesive semantic boundaries. Similarly, in the ResNet-50 example, Guided Backprop results in visibly entangled concepts. These qualitative results align with our quantitative findings, confirming that while our default choice of Input × Gradient provides reliable and distinct conceptual disentanglement, the SPICE framework is broadly applicable across various attribution techniques.

\paragraph{Compatibility with Recent Attribution Methods (LeGrad).}
To further demonstrate that SPICE is agnostic to the choice of attribution backbone, we additionally evaluate it with LeGrad~\cite{bousselham2024legrad}, a recent Transformer-tailored attribution method. We adapt LeGrad to produce layer-conditional, feature-wise scores aligned with SPICE's per-neuron footprint formulation. Table~\ref{appx:tab:legrad} reports separability at ViT-B/16 blocks B2, B6, and B11. SPICE retains the same depth-wise trend with LeGrad as with Input $\times$ Gradient, exhibiting only minor variance from the default. We leave the integration of OMENN~\cite{wrobel2024omenn} for future work, as no public code is currently available.

\begin{table}[h]
\centering\small
\caption{Compatibility of SPICE with different attribution backbones at ViT-B/16 (separability $\uparrow$).}
\label{appx:tab:legrad}
\setlength{\tabcolsep}{6pt}
\begin{tabular}{lccc}
\toprule
Attribution & B2 & B6 & B11 \\
\midrule
\rowcolor{gray!15}
Input $\times$ Gradient (default) & 1.137 & 1.240 & 1.577 \\
LeGrad (layer-conditional)        & 1.111 & 1.196 & 1.493 \\
\bottomrule
\end{tabular}
\end{table}

\subsection{Threshold ($\tau$) Sensitivity}
\label{appx:subsec:tau_sens}

\begin{figure}[!b]
    \centering
    \includegraphics[width=0.9\textwidth]{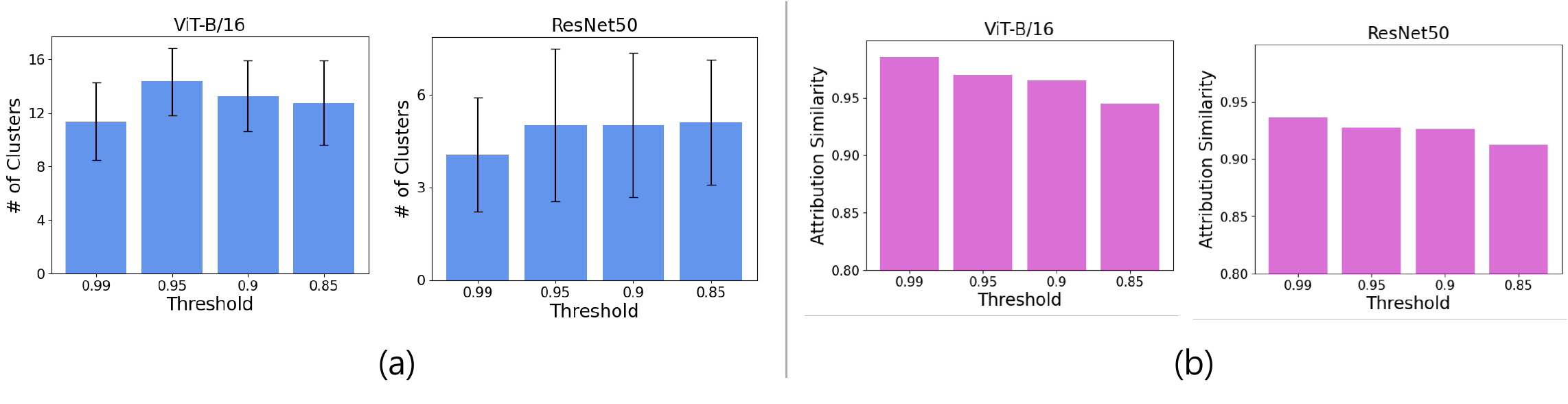}
    \caption{Sensitivity analysis on the threshold of the last layer/block of ViT-B/16 and ResNet-50. Left \textbf{(a)} shows the number of concept clusters, and right \textbf{(b)} shows the attribution similarity of concept clusters at different thresholds.}
    \label{appx:fig:appx_sens}
\end{figure}

In our proposed method, we construct flexible concept clusters tailored to each model and dataset by selecting an appropriate threshold. Specifically, we adopt the larger value between the second peak of the similarity distribution (estimated using KDE) and the score at the top-p$\%$ quantile as the threshold. To examine the sensitivity of this hyperparameter, we varied the value of p and measured its impact on both the number of clusters formed and the within-cluster similarity. The results are shown in Figure~\ref{appx:fig:appx_sens}. We conducted experiments with p = [0.99, 0.95, 0.9, 0.85] on ResNet50 and ViT-Base-16 models. As shown in Figure~\ref{appx:fig:appx_sens}-(a), ViT consistently produced more concept clusters on average compared to ResNet50, which aligns with prior findings that polysemantic neurons are more prevalent and complex in transformer architectures. Interestingly, as the threshold became looser, the number of clusters decreased, whereas exhibited the fewest clusters at p=0.99. This behavior arises because, at such a strict threshold, some clusters fail to form due to the distances between samples in the representation space being larger than the cutoff. Figure~\ref{appx:fig:appx_sens}-(b) reports the within-cluster similarity scores across thresholds, all of which remain significantly high (\textgreater 0.9). A slight downward trend is observed as the threshold decreases, indicating that looser thresholds admit some noisier samples. Taken together, these results show that p=0.95 provides the best trade-off, reliably producing meaningful clusters across architectures while maintaining the highest within-cluster similarity.

\subsection{Effect of Normalization}
\label{appx:subsec:abl-normalize}

We investigate the effect of normalizing the footprint matrix \(\mathcal{A}_u \in \mathbb{R}^{|\mathcal{D}_{u^l}| \times d_{l-1}}\) along the neuron dimension before clustering, following the definition used in the main paper. Concretely, for each predecessor-neuron column \(j \in \{1,\dots,d_{l-1}\}\) we apply feature-wise standardization,
\begin{equation}
\tilde{\mathcal{A}}_u[i,j] \;=\; \frac{\mathcal{A}_u[i,j] - \mu_j}{\sigma_j},
\qquad
\mu_j = \frac{1}{|\mathcal{D}_{u^l}|}\sum_{i} \mathcal{A}_u[i,j],
\qquad
\sigma_j^2 = \frac{1}{|\mathcal{D}_{u^l}|}\sum_{i} \bigl(\mathcal{A}_u[i,j] - \mu_j\bigr)^2,
\end{equation}
so that each neuron dimension has zero mean and unit variance across the sampled footprints, analogous to the standardization commonly used in time-series analysis. Raw footprint columns often exhibit significant heterogeneity in scale and variance across neurons; without this adjustment, dimensions with inherently larger magnitudes disproportionately dominate the clustering process, obscuring the semantic structures we aim to identify. By normalizing along the neuron dimension, we ensure that the clustering is driven by the relative distribution of importance (the semantic pattern) rather than absolute magnitude biases. As reported in Table~\ref{appx:tab:ablation_norm}, this normalization step consistently improves separability across both CNN and Transformer architectures and across early/late layers, corroborating its role as a core component of SPICE.

\begin{table}[h]
\centering
\caption{Model Layer Attribution Normalization and Separability}
\label{appx:tab:ablation_norm}
\begin{tabular}{l|l|c|c}
\toprule
\textbf{Model} & \textbf{Layer} & \textbf{Normalization} & \textbf{Separability} \\
\midrule
ResNet-50 & L2.1 &  & 1.058 \\
ResNet-50 & L2.1 & \checkmark & \textbf{1.092} \\
\midrule
ResNet-50 & L4.2 &  & 1.017 \\
ResNet-50 & L4.2 & \checkmark & \textbf{1.293} \\
\midrule
ViT-B & B2 &  & 1.065 \\
ViT-B & B2 & \checkmark & \textbf{1.137} \\
\midrule
ViT-B & B11 &  & 1.507 \\
ViT-B & B11 & \checkmark & \textbf{1.577} \\
\bottomrule
\end{tabular}
\end{table}

\subsection{Stepwise Ablation: From PURE to SPICE}
\label{appx:subsec:pure_to_spice}
To clarify the individual contribution of each SPICE component over the PURE baseline, we conduct a stepwise ablation at ViT-B/16 block~11. Starting from PURE-style fixed $K\!=\!2$ clustering on CRP attribution, we sequentially (i) replace CRP with Input $\times$ Gradient (IxG), (ii) replace fixed $K$ with our adaptive cluster discovery, and (iii) add per-dimension attribution normalization, recovering the full SPICE configuration. As shown in Table~\ref{appx:tab:pure_to_spice}~(left), the adaptive threshold and the normalization step both contribute substantial gains, with normalization helping mitigate feature-space anisotropy as discussed in Section~\ref{sec:method} of the main paper. We also directly compare against fixed-$K$ baselines under our pipeline in Table~\ref{appx:tab:pure_to_spice}~(right), confirming that no choice of constant $K \in \{2,3,4\}$ matches the adaptive variant.

\begin{table}[h]
\centering\footnotesize
\caption{Ablation at ViT-B/16 block~11 (separability $\uparrow$). \textbf{Left}: stepwise PURE$\to$SPICE. \textbf{Right}: fixed-$K$ baselines vs.\ adaptive cluster discovery.}
\label{appx:tab:pure_to_spice}
\setlength{\tabcolsep}{4pt}
\begin{minipage}[t]{0.50\linewidth}
\centering
\begin{tabular}{lc}
\toprule
Configuration & Sep. \\
\midrule
$K\!=\!2$, CRP ($\approx$PURE) & 1.036 \\
$K\!=\!2$, IxG                 & 1.027 \\
Adaptive $K$, IxG              & 1.308 \\
\rowcolor{gray!15}
\quad + Normalization ($=$ SPICE) & \textbf{1.577} \\
\bottomrule
\end{tabular}
\end{minipage}%
\begin{minipage}[t]{0.50\linewidth}
\centering
\begin{tabular}{lc}
\toprule
Configuration & Sep. \\
\midrule
Fixed $K\!=\!2$ & 1.027 \\
Fixed $K\!=\!3$ & 1.025 \\
Fixed $K\!=\!4$ & 1.026 \\
\rowcolor{gray!15}
Adaptive (Ours) & \textbf{1.577} \\
\bottomrule
\end{tabular}
\end{minipage}
\end{table}

\subsection{Choice of Clustering Method}
We compare the effectiveness of three different clustering algorithms in generating well-separated concepts: K-Means (our default), ISODATA~\cite{ball1965isodata}, and HDBSCAN~\cite{campello2013density}. ISODATA adaptively splits and merges samples based on the inter-cluster distance and the intra-cluster variance. To enable a fair comparison while preserving the spirit of our attribution-based clustering method, we adapt ISODATA by performing cluster merging based on our proposed score and cluster splitting based on the silhouette score. We set the splitting threshold to the critical value at which the behavior begins to deviate from our method, and used this setting to examine whether the direction of changes induced by the splitting logic is reasonable.

\begin{figure}
    \centering
    \includegraphics[width=1\linewidth]{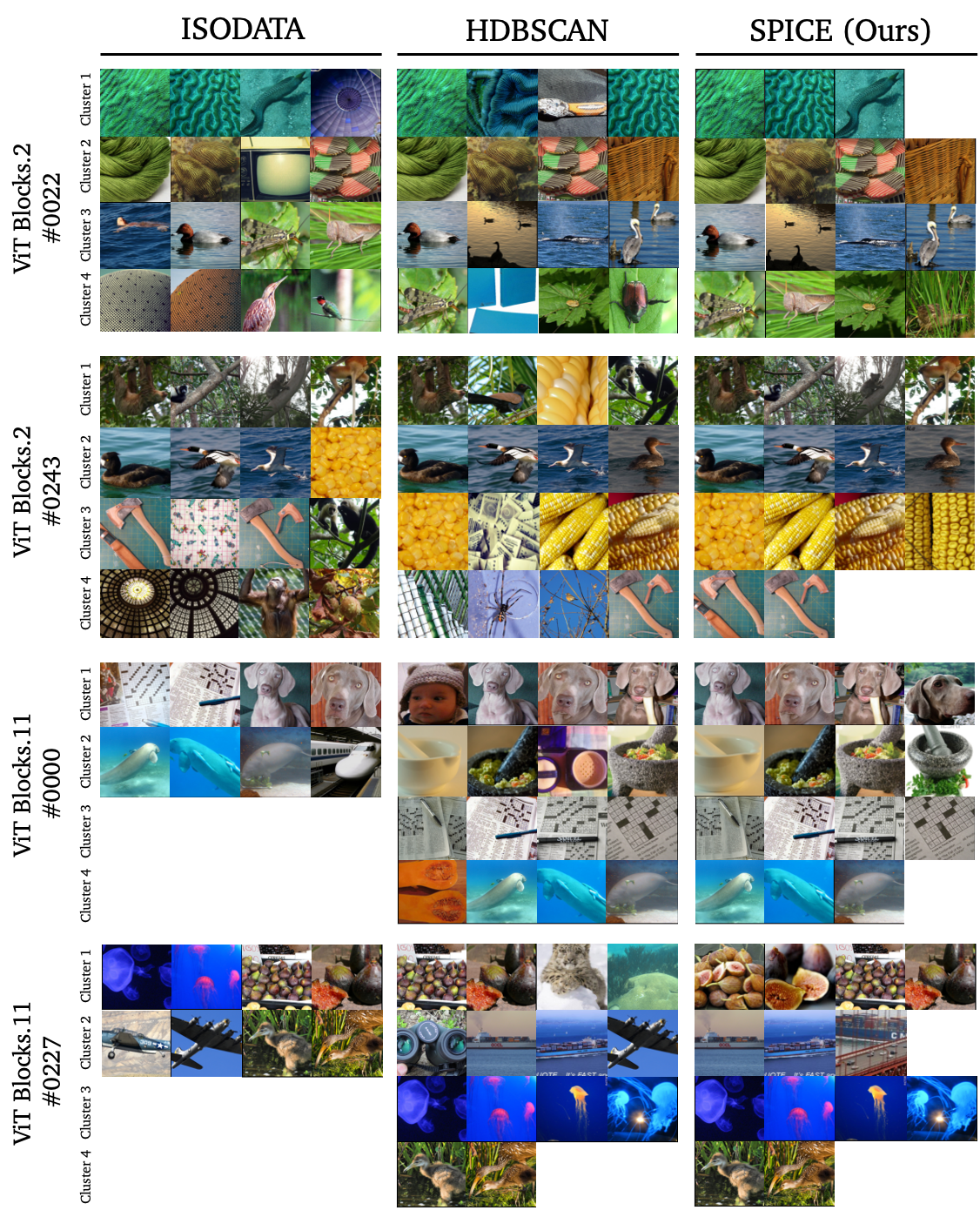}
    \caption{Ablation study of the clustering components in our method, comparing the use of ISODATA, HDBSCAN, and K-Means for computing cluster means. The last column shows the default version of our algorithm, which employs iterative K-Means.}
    \label{appx:fig:clustering}
\end{figure}

As illustrated in Figure~\ref{appx:fig:clustering}, our proposed method (SPICE) demonstrates superior performance in isolating semantically consistent concepts across various network depths. Unlike ISODATA and HDBSCAN, which often produce noisy or entangled clusters, our approach effectively disentangles distinct visual concepts within a single neuron. Notably, SPICE maintains high semantic coherence for each identified cluster while successfully capturing a diverse range of features, demonstrating its robustness across different layers of the model.

\subsection{Correlation between Cluster Count and Separability}
\label{appx:subsec:correlation}
To validate our clustering algorithm and address potential concerns regarding over-segmentation, we analyzed the relationship between the number of discovered concepts ($k$) and their separability scores. We observed no significant correlation (Spearman $\rho = -0.041, p = 0.682$), alongside negligible differences across groups with varying $k$ (ANOVA $p=0.717$). This lack of correlation is crucial: it implies that SPICE does not artificially over-cluster features to inflate performance. Instead, the consistent separability across varying $k$ confirms that our method adaptively discovers the optimal number of concepts based on the neuron's intrinsic polysemanticity, maintaining high validity for both simple and highly complex neurons.

\begin{figure}
    \centering
    \includegraphics[width=\textwidth]{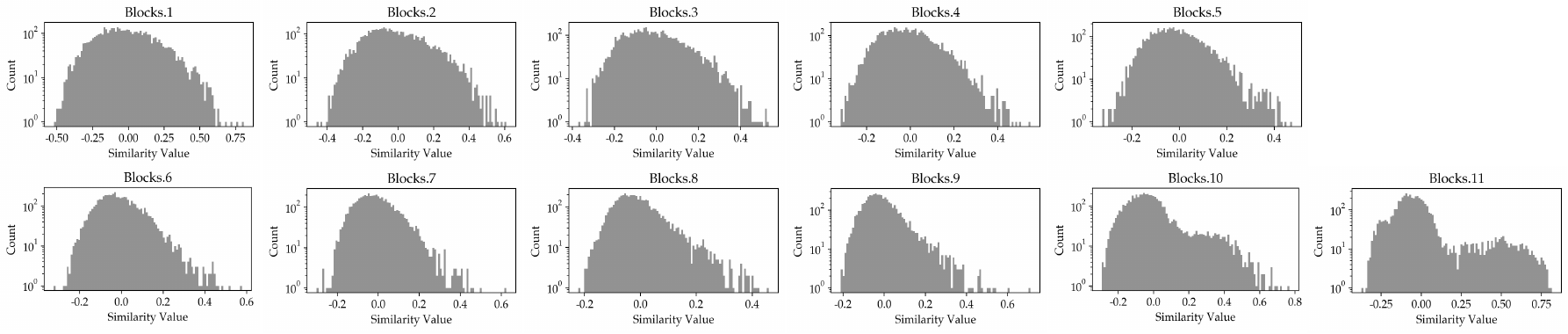}
    \caption{Changes in the distribution of attribution similarity across ViT-B’s blocks, shown from the first block (top-left) to the last block (bottom-right).}
    \label{appx:fig:appendix_distributions}
\end{figure}

\begin{figure}
    \centering
    \includegraphics[width=\textwidth]{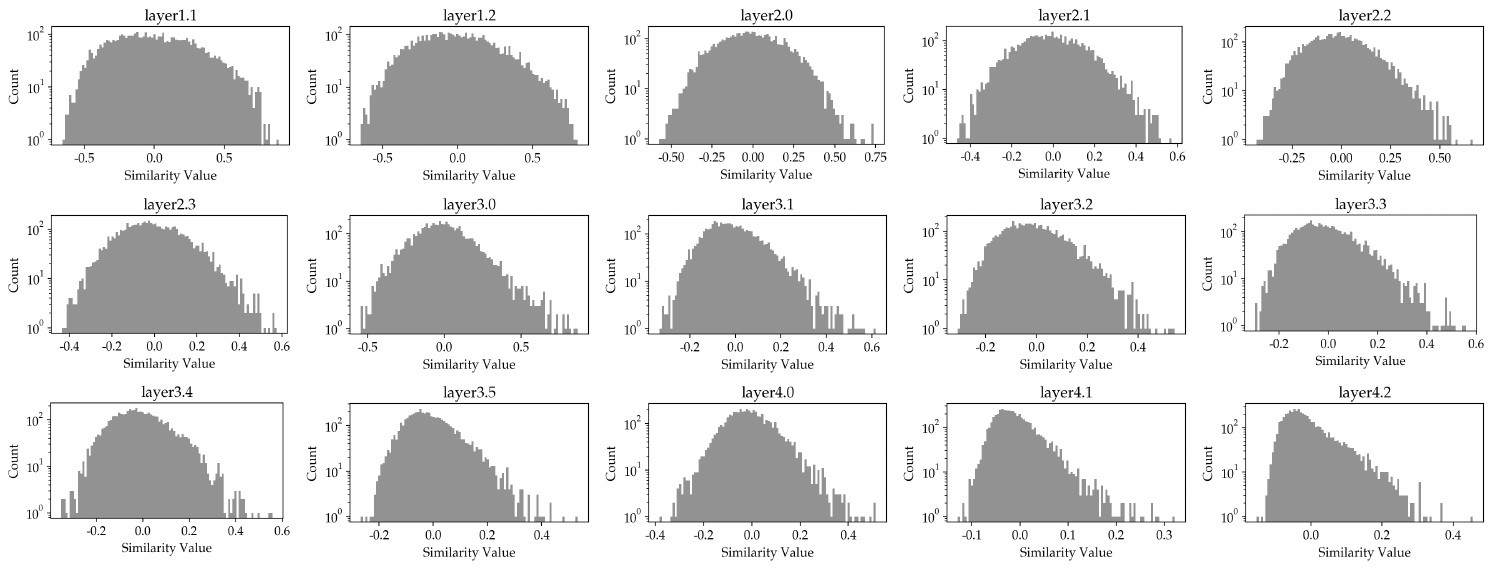}
    \caption{Changes in the distribution of attribution similarity across ResNet-50’s layers, shown from the first layer (top-left) to the last layer (bottom-right).}
    \label{appx:fig:appendix_distributions_resnet}
\end{figure}

\section{Structural Evolution of Concepts Across Network Depth}
\label{appx:sec:depth_evolution}

In this section, we provide a detailed analysis of how the structure of attribution similarity evolves across layers, justifying the necessity of our dual-criterion thresholding strategy defined in Section~\ref{sec:method} in the main text. Figures~\ref{appx:fig:appendix_distributions} and \ref{appx:fig:appendix_distributions_resnet} visualize these distributions for ViT-Base and ResNet-50, respectively, while Table~\ref{appx:tab:vit_vs_resnet_dip} quantifies their unimodality using Hartigan's Dip Test~\cite{hartigan1985dip}. As shown in Figure~\ref{appx:fig:appendix_distributions}, ViT-Base exhibits a clear transition. Early layers are robustly unimodal, but deeper layers (10--11) begin to develop a bimodal structure for a significant subset of neurons, forming a distinct secondary peak in the high-similarity region. This is statistically captured by the Dip Test $p$-values in Table~\ref{appx:tab:vit_vs_resnet_dip}; while the averages drop to 0.60 and 0.28, the accompanying high variance ($\pm 0.42, \pm 0.40$) is highly indicative. Mathematically, this broad distribution implies that rather than every neuron becoming universally multimodal, a substantial subpopulation of neurons clearly diverges from strict unimodality, signaling the emergence of highly compressed, polysemantic concepts. In contrast, ResNet-50 remains consistently unimodal ($p > 0.9$) across all depths. While deeper layers in ResNet become right-skewed, they do not form a separable secondary mode.

Although ResNet-50 stays unimodal under the Dip Test, its attribution-similarity distribution is not flat with depth: skewness rises monotonically from $0.30$ at Layer~1 to $1.40$ at Layer~4, i.e., the high-similarity tail becomes increasingly heavy in deeper blocks (Figure~\ref{appx:fig:appendix_distributions_resnet}). Hence the qualitative trend of growing attribution similarity with depth is shared by CNNs, even though they never form a separable secondary mode as ViTs do. Quantifying the multimodal subpopulation directly, $43\%$ of neurons at ViT-B/16 block~11 are \emph{significantly} multimodal under the Dip Test at $p<0.05$, supporting our reading of Table~\ref{appx:tab:vit_vs_resnet_dip} as evidence for a substantial subpopulation rather than a layer-wide shift.

These observations highlight why a single thresholding method is insufficient. For bimodal distributions (e.g., deep ViT layers), a fixed percentile ignores the true distribution of the data. Instead, the local maximum (the second peak) identified by KDE pinpoints the exact center of a semantically cohesive cluster. Since we measure cosine similarity, leveraging this high-similarity peak as the threshold ensures that only the most strongly related pathways are grouped, leaving out the noisy middle ground. Conversely, for unimodal distributions (e.g., ResNet or early ViT), the absence of a secondary peak makes KDE thresholding invalid. To dynamically accommodate both structural behaviors, we set the threshold as $\tau = \max(\text{95th percentile}, \text{KDE 2nd peak})$. This selectively applies strict, peak-driven thresholding for highly polysemantic layers while safely defaulting to a 95th-percentile baseline for unimodal spaces.

\begin{table}[h]
    \centering
    \caption{Comparison of distribution statistics between ViT-Base and ResNet-50. ResNet-50 shows consistently high Hartigan’s Dip Test $p$-values ($> 0.9$), indicating stable unimodal distributions across all depths. In contrast, ViT-Base exhibits a precipitous drop in mean $p$-values accompanied by exceptionally high variance in its final layers (\textbf{bold}). This broad distribution indicates the emergence of a substantial subpopulation of highly multimodal (polysemantic) neurons, rather than a universal shift across the entire layer. Depth is measured in a block-wise or layer-wise manner, depending on the model architecture.}
    \label{appx:tab:vit_vs_resnet_dip} 
    \begin{tabular}{lcc}
        \toprule \hline
        \textbf{Depth} & \textbf{ViT-Base} & \textbf{ResNet50} \\
        \midrule
        Layer 1  & $0.98 \pm 0.06$ & $0.95 \pm 0.17$ \\
        Layer 2  & $0.98 \pm 0.07$ & $0.97 \pm 0.13$ \\
        Layer 3  & $0.98 \pm 0.09$ & $0.96 \pm 0.11$ \\
        Layer 4  & $0.98 \pm 0.06$ & $0.97 \pm 0.12$ \\
        Layer 5  & $0.97 \pm 0.09$ & $0.97 \pm 0.08$ \\
        Layer 6  & $0.98 \pm 0.06$ & $0.97 \pm 0.08$ \\
        Layer 7  & $0.99 \pm 0.06$ & $0.97 \pm 0.11$ \\
        Layer 8  & $0.98 \pm 0.08$ & $0.98 \pm 0.08$ \\
        Layer 9  & $0.94 \pm 0.19$ & $0.97 \pm 0.09$ \\
        Layer 10 & \textbf{0.60 $\pm$ 0.42} & $0.97 \pm 0.09$ \\
        Layer 11 & \textbf{0.28 $\pm$ 0.40} & $0.97 \pm 0.10$ \\
        Layer 12 & ---             & $0.97 \pm 0.10$ \\
        Layer 13 & ---             & $0.96 \pm 0.16$ \\
        Layer 14 & ---             & $0.97 \pm 0.12$ \\
        Layer 15 & ---             & $0.98 \pm 0.09$ \\
        \hline \bottomrule 
    \end{tabular}
\end{table}

\section{Running Time Analysis}
In this section, we provide a formal derivation of the theoretical worst-case time complexity of our iterative clustering algorithm and contrast it with the empirical running time observed in practice.

\subsection{Theoretical Worst-Case Complexity}
Each iteration of the algorithm involves K-Means clustering, which scales as $\mathcal{O}(k \cdot N \cdot d)$, and pairwise similarity validation, which incurs a cost of $\mathcal{O}(N^2)$. Treating the feature dimension $d$ as a constant, the per-iteration computational cost is dominated by the $\mathcal{O}(N^2)$ validation step. The theoretical worst-case scenario occurs if the algorithm fails to satisfy the cohesion threshold at every step, causing the number of search clusters $k$ to increment linearly up to $N$ without reducing the working set. Summing the $\mathcal{O}(N^2)$ cost across $\mathcal{O}(N)$ iterations results in a theoretical upper bound of $\mathcal{O}(N^3)$.

\subsection{Empirical Analysis}

Despite the cubic theoretical bound, the worst-case scenario (complete rejection of clusters) rarely occurs in real-world experiments. This is primarily because the input samples are derived from the high activations of a single neuron, which inherently exhibit strong semantic coherence. Since these samples tend to cluster tightly around distinct concepts in the feature space, the algorithm can rapidly identify and accept valid groups satisfying the similarity threshold.
This prevents the pathological case of repeated rejections and allows the algorithm to efficiently prune the search space.

To verify this practical scalability, we measured the wall-clock time across varying sample sizes $N$. Table~\ref{appx:tab:running_time} reports the observed time and a complexity check metric, $T / (N \log_{10} N)$. As shown in the table, this metric stabilizes around $4.0 \times 10^{-4}$ for $N \ge 1,000$. This stability confirms that, aided by the inherent structure of neuron activations, the algorithm follows an effective complexity of $\mathcal{O}(N \log N)$.

\begin{table}[h!]
\centering
\caption{Computational Efficiency of Iterative Divisive Clustering}
\label{appx:tab:running_time}
\begin{tabular}{>{\centering\arraybackslash}p{2cm} >{\centering\arraybackslash}p{2cm} >{\centering\arraybackslash}p{2.8cm}}
    \toprule
    \textbf{Sample Size ($N$)} & \textbf{Observed Time (s)} & \textbf{Complexity Check} ($\text{Time} / (N \log_{10} N)$) \\
    \midrule
    100 & 1.45 & 0.003149 \\
    200 & 4.63 & 0.004369 \\
    500 & 1.89 & 0.000608 \\
    1,000 & 3.01 & 0.000436 \\
    2,000 & 5.94 & 0.000391 \\
    5,000 & 17.71 & 0.000416 \\
    \bottomrule
\end{tabular}
\end{table}

\begin{figure}[!h]
    \centering
    \includegraphics[width=\textwidth]{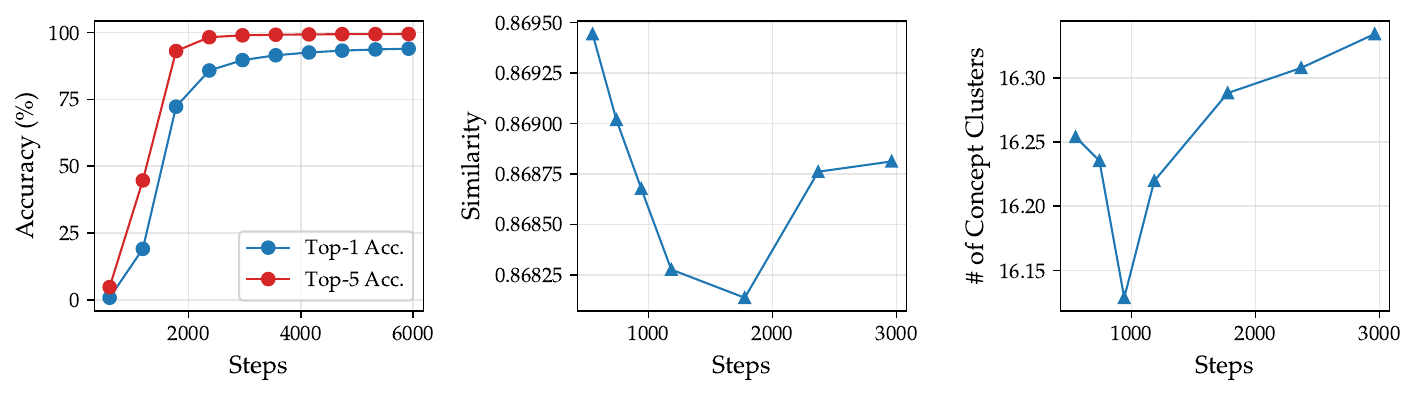}
    \caption{Training accuracy plotted against training steps during transfer learning of ViT-L. The left panel displays Top-1 and Top-5 classification accuracy. The middle and right panels illustrate the attribution similarity and the average number of concept clusters, respectively.}
    \label{appx:fig:dino_stat}
\end{figure}

\section{Training Dynamics and Transfer Learning}

\begin{figure}[t]
    \centering
    \includegraphics[width=0.4\columnwidth]{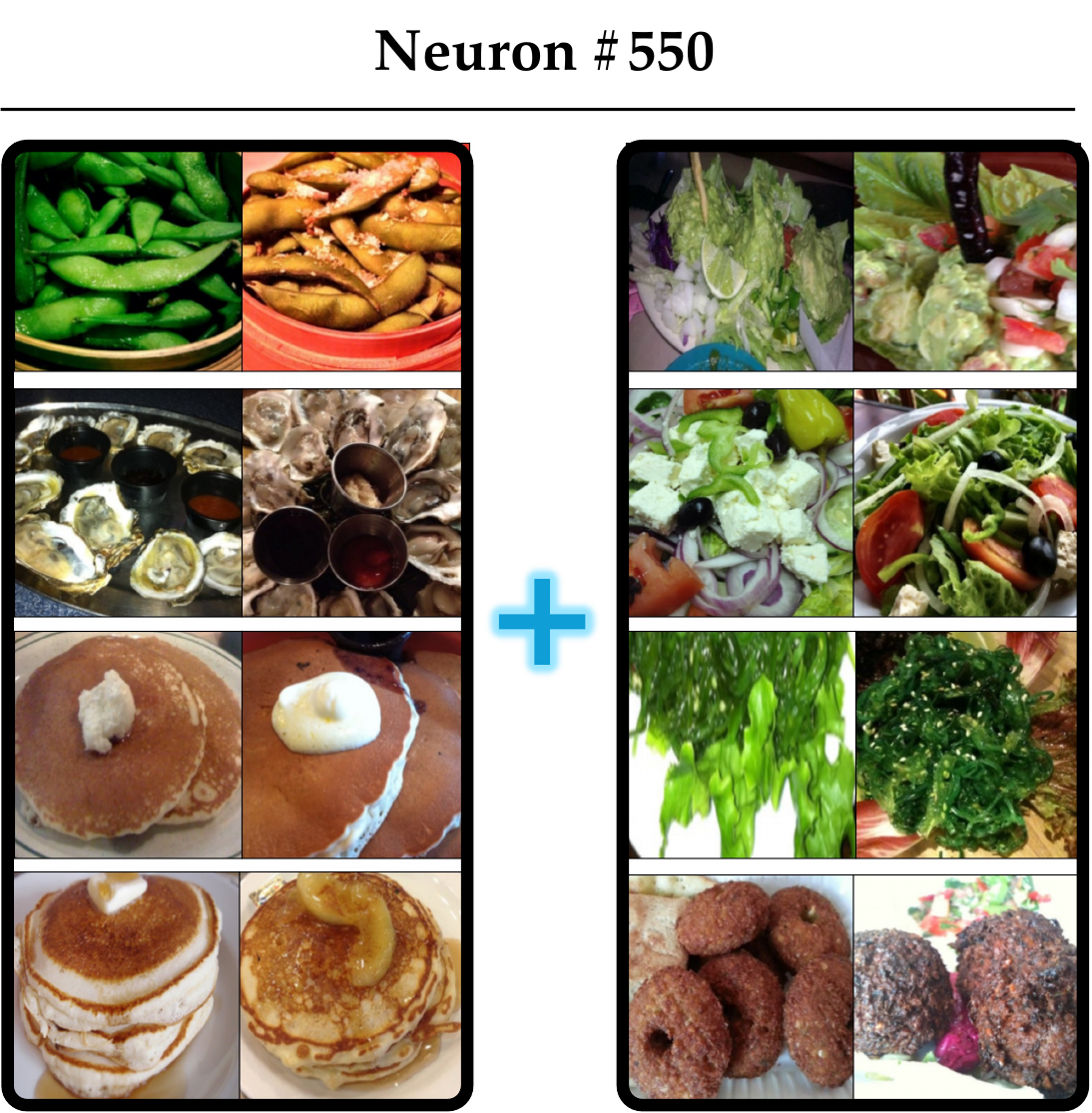}
    \caption{Visualization of concept expansion in Neuron~\#550. The left panel shows the top-activating concepts at initialization, while the right panel shows the new concepts added after fine-tuning.}
    \label{appx:fig:dino_case}
\end{figure}
To investigate the evolution of feature representations during transfer learning, we fine-tuned a ViT-Large (ViT-L) model pre-trained on the LVD-1689M dataset using the DINOv3~\cite{simeoni2025dinov3} framework. The downstream task was performed on the Food-101 dataset~\cite{bossard2014food}. To ensure parameter efficiency and focus on the adaptation of high-level semantic features, we updated only the Multi-Layer Perceptron (MLP) weights of the final Transformer block and the linear classification head. The model was trained for 6 epochs. To capture the rapid dynamic changes characteristic of the early training phase, metrics were logged every 50 steps during the first two epochs. 

We begin our analysis of transfer-learning dynamics by examining how individual feature representations evolve during fine-tuning. Specifically, we first focus on neuron-level semantic changes to understand how pre-trained visual concepts adapt to the new task domain. Figure~\ref{appx:fig:dino_case} presents a representative example—Neuron \#550—which exhibited the most extensive concept expansion during fine-tuning. Initially sensitive to visual attributes such as the green color of \textit{Edamame} or the brown, textured appearance of \textit{Pancakes}, this neuron gradually broadened its activations to include semantically distinct yet visually similar categories such as \textit{Greek Salad}, \textit{Guacamole}, and \textit{Falafel}. This case illustrates how transfer learning often repurposes existing filters to cover new visual clusters rather than forming entirely new features.

The global training behaviors are summarized in Figure~\ref{appx:fig:dino_stat}, which reports accuracy metrics, attribution similarity, and the evolution of concept clusters. When we assigned representative concepts to each neuron based on the majority class labels of the activating samples, we found that, across the 1,024 neurons in the targeted MLP layer, 284 neurons (approximately 28\%) preserved their original concept clusters between the initial and final checkpoints. This substantial stability indicates that DINOv3 pre-training yields robust visual features that remain largely suitable for the downstream task. Given that the feature dimensionality (1,024) far exceeds the number of target classes (101), the model retains a large portion of its pre-trained representational space rather than reorganizing it wholesale.

\section{Robustness across Probing Datasets}
\label{appx:sec:probing_dataset}
Our main separability evaluation uses ImageNet validation as the probing dataset. To check whether SPICE's behavior generalizes beyond this choice, and to disentangle pretraining-distribution effects from any probing-dataset bias, we additionally evaluate ViT-B/16 and CLIP ViT-B/16 at block~11 on Places365~\cite{zhou2017places} and YFCC100M~\cite{thomee2016yfcc100m}. As shown in Table~\ref{appx:tab:probing_dataset}, the relative ranking shifts in a manner consistent with each model's pretraining: ViT-B/16 leads on ImageNet, CLIP ViT-B/16 leads on YFCC100M (near its pretraining distribution), and the two models become comparable on Places365 (out-of-distribution for both). Crucially, SPICE yields consistently high separability across all three probing distributions; the absolute differences across columns reflect pretraining and dataset choice rather than instability of our method.

\begin{table}[h]
\centering\small
\caption{Separability of SPICE at the last block of ViT-B/16 and CLIP ViT-B/16 across three probing datasets.}
\label{appx:tab:probing_dataset}
\setlength{\tabcolsep}{6pt}
\begin{tabular}{lccc}
\toprule
Model & ImageNet & YFCC100M & Places365 \\
\midrule
ViT-B/16 B11      & \textbf{1.577}\,{\tiny$\pm$0.094} & 1.359\,{\tiny$\pm$0.133} & 1.415\,{\tiny$\pm$0.123} \\
CLIP ViT-B/16 B11 & 1.410\,{\tiny$\pm$0.102} & \textbf{1.396}\,{\tiny$\pm$0.103} & \textbf{1.423}\,{\tiny$\pm$0.100} \\
\bottomrule
\end{tabular}
\end{table}

\section{Diverse Models and Layers Analysis}
\label{appx:label:diverse_analysis}

\begin{figure}[!t]
    \centering
    \includegraphics[width=0.87\textwidth]{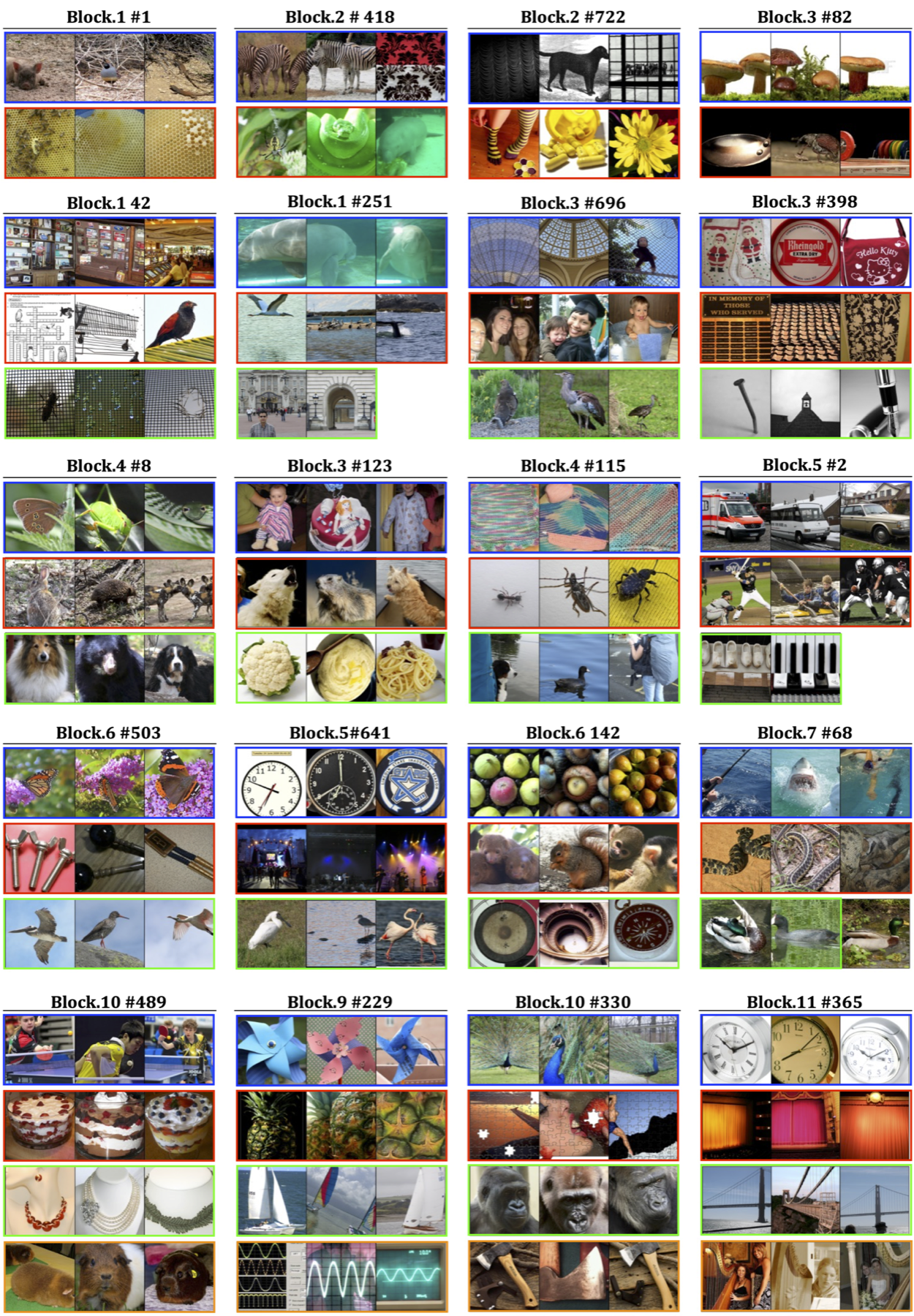}
    \caption{Examples of polysemantic neurons of ViT-B with different blocks. Each row in a neuron represents a distinct concept cluster.}
    \label{appx:fig:spice_vit}
\end{figure}

\begin{figure}[t]
    \centering
    \includegraphics[width=0.87\textwidth]{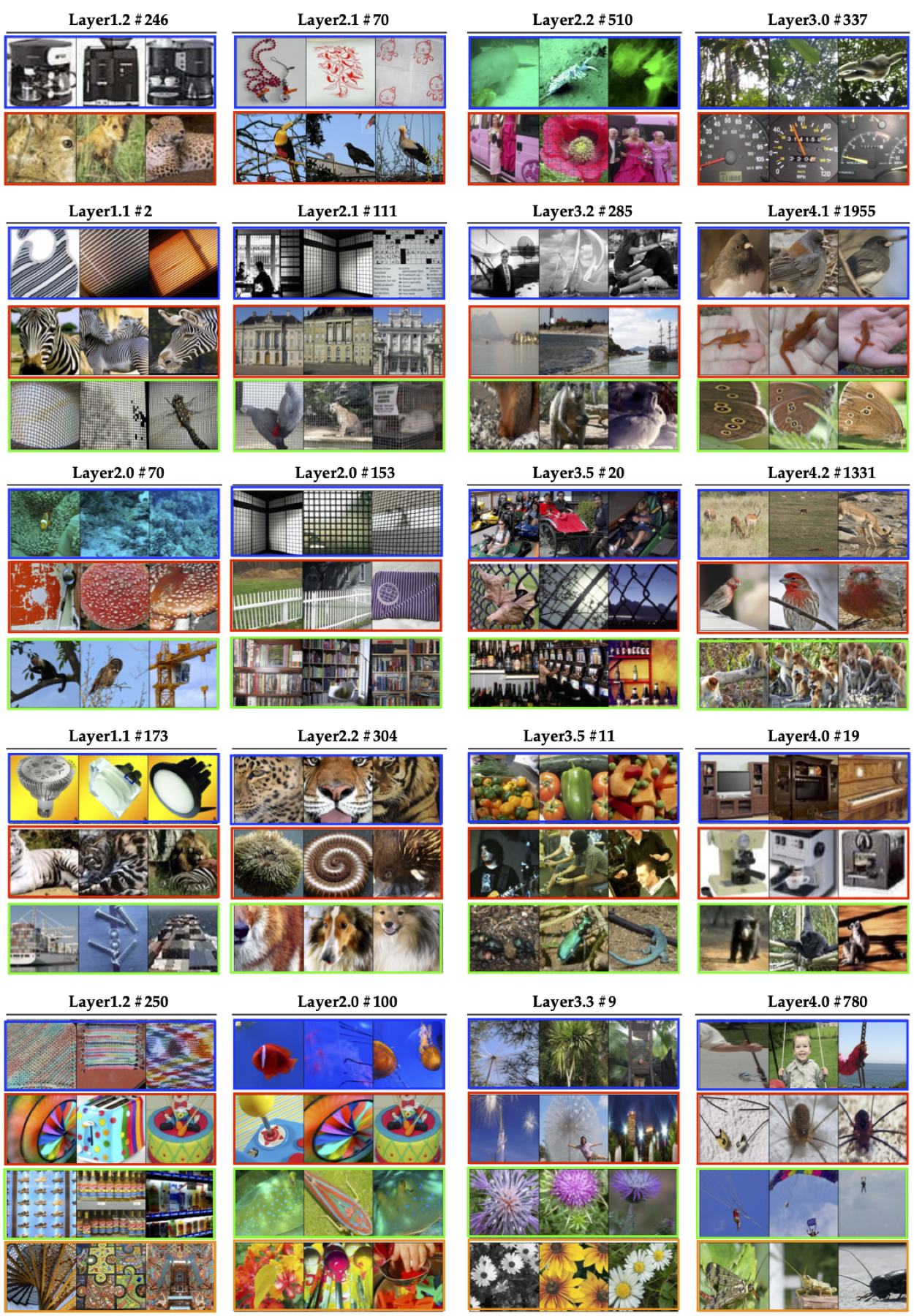}
    \caption{Examples of polysemantic neurons of ResNet-50 with different blocks. Each row in a neuron represents a distinct concept cluster.}
    \label{appx:fig:spice_resnet}
\end{figure}

\begin{figure}[t]
    \centering
    \includegraphics[width=0.87\textwidth]{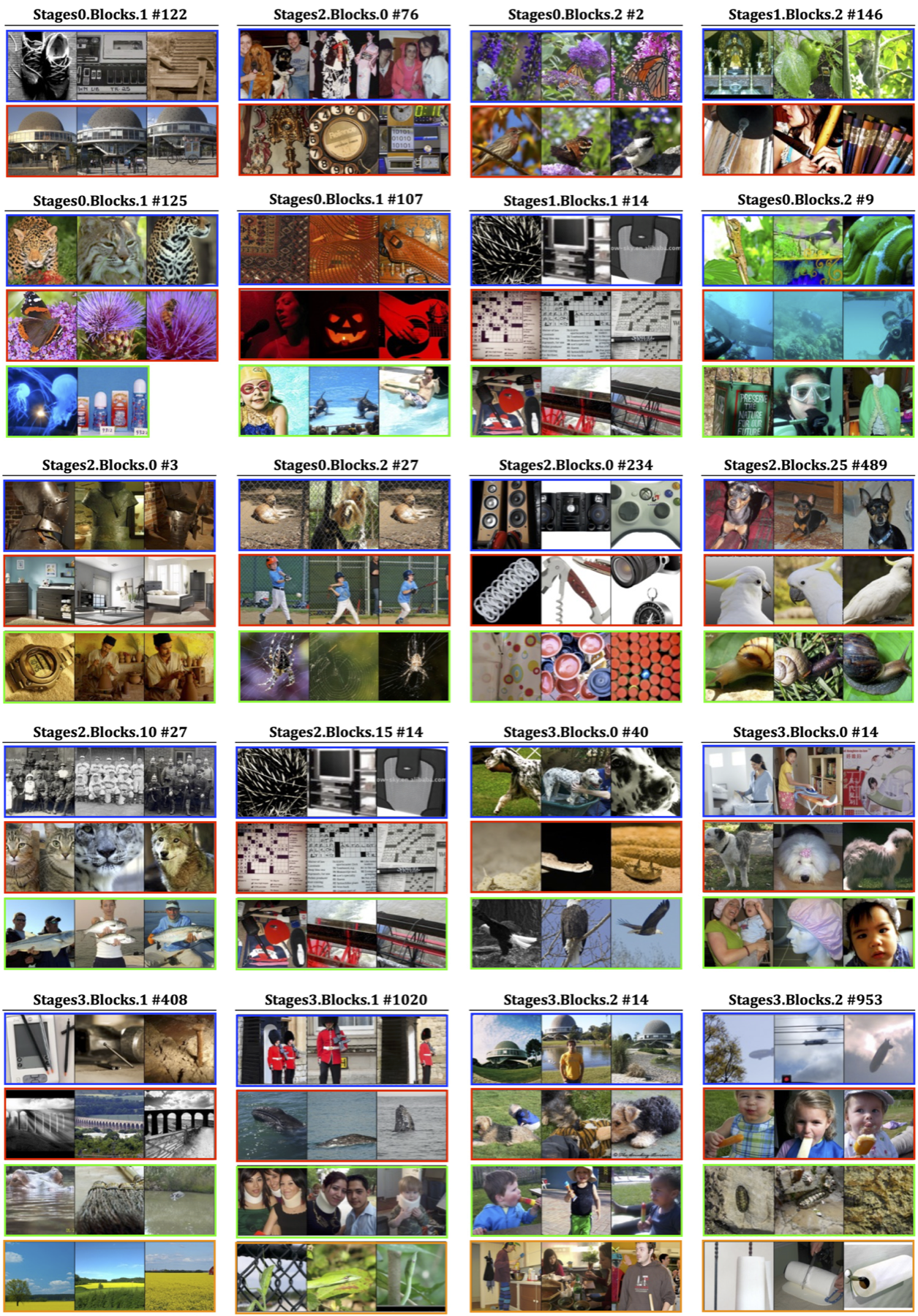}
    \caption{Examples of polysemantic neurons of ConvNeXt-B with different blocks. Each row in a neuron represents a distinct concept cluster.}
    \label{appx:fig:spice_convnext}
\end{figure}

We extend the analysis to ensure our findings are not architecture-specific. Figures~\ref{appx:fig:spice_vit}, \ref{appx:fig:spice_convnext}, and \ref{appx:fig:spice_resnet} demonstrate that our method produces semantically coherent concepts across different feature abstraction levels (early, middle, and late layers). The consistent improvement observed across this diverse set of backbones confirms the general applicability of our framework in enhancing interpretability for modern deep learning architectures. For transparency, the neurons displayed in Figure~\ref{fig:polysemantic_neuron_dissection} of the main paper and in the qualitative figures of this appendix were sampled uniformly at random from the corresponding layers (i.e., they are not cherry-picked).

\clearpage

\end{document}